\documentclass{article}

\usepackage[nonatbib,preprint]{neurips_2026}

\usepackage[utf8]{inputenc} 
\usepackage[T1]{fontenc}    
\usepackage{hyperref}       
\usepackage{url}            
\usepackage{booktabs}       
\usepackage{amsfonts}       
\usepackage{nicefrac}       
\usepackage{microtype}      
\usepackage{xcolor}         

\usepackage{subcaption}
\usepackage{wrapfig}
\usepackage{dsfont}
\usepackage[makeroom]{cancel}
\usepackage{amsmath}
\usepackage{amssymb}
\usepackage{mathtools}
\usepackage{amsthm}
\usepackage{multirow}
\usepackage{tikz}
\usetikzlibrary{arrows.meta, positioning, quotes}

\let\citep\cite
\let\citet\cite
\newtheorem{theorem}{Theorem}[section]
\newtheorem{proposition}[theorem]{Proposition}

\newtheorem{corollary}[theorem]{Corollary}
\newtheorem{definition}[theorem]{Definition}

\title{Transformer as an Euler Discretization of Score-based Variational Flow}

\author{%
  Huadong Liao \\
  \texttt{naturomics.liao@gmail.com} \\
}

\begin{document}
\maketitle

\begin{abstract}
Despite the Transformer's dominance across machine learning, its architecture remains largely heuristic and lacks a unified theoretical foundation.
We introduce Score-based Variational Flow (SVFlow), a continuous‑time dynamical system for representation learning in which
the state evolves according to a variational posterior–weighted average of conditional log-likelihood scores,
and provide a principled basis for regularization through variational consistency.
We show that forward Euler discretization of spherical SVFlow exactly recovers the Transformer architecture.
Multi‑head attention approximates SVFlow vector field via a vMF kernel‑smoothed posterior, while MoE/FFN approximates it in a relaxed network-based way,
and the residual‑normalization block implements a relaxed retraction that maintains spherical geometry.
This unification explains why attention trains stably without explicit regularization while MoE requires auxiliary balancing losses.
Experiments on pre‑trained language models with prefix shuffling show that SVFlow‑induced metrics correlate with task performance,
reveal depth‑dependent sensitivity, and reflect the intrinsic dynamics of attention.
\end{abstract}

\section{Introduction}
\label{intro}
The Transformer~\citep{vaswani2017attention} has become the cornerstone of modern machine learning, yet its design remains fundamentally heuristic.
Recent efforts have retroactively explained its components: attention as kernel methods~\citep{tsai2019transformer} or gradient descent~\citep{agarwal2026gradient},
residual connections as neural ODEs~\citep{chen2018neural}, and normalization as geometric projection~\citep{gupta2025layernorm}.
While insightful, these perspectives are fragmented, leaving many basic phenomena unexplained. For instance,
multi‑head attention learns stable representations without explicit regularization,
whereas mixture‑of‑experts (MoE) layers require auxiliary balancing losses to prevent collapse~\citep{shazeer2017,fedus2022switch}.

To unify the fragmented perspectives of previous work, we propose \textbf{Score-based Variational Flow (SVFlow)}, a continuous‑time dynamical system for representation learning.
In SVFlow, the state evolves according to the variational posterior–weighted average of conditional score functions, combining dynamic filtering with likelihood ascent.
The framework provides a principled basis for regularization through a variational objective,
and formalizes a gradient‑dynamics analysis that reveals the implicit self‑regularization of the attention posterior.
It also naturally induces metrics, including marginal likelihood, divergence, and concentrations, for probing model behavior.

SVFlow itself is agnostic to the parametric family. Instantiating it on the spherical manifold yields the natural geometric setting for Transformers.
The forward Euler discretization of such spherical SVFlow then explicitly links each Transformer component to its counterpart:
multi‑head attention approximates SVFlow vector field via a von Mises-Fisher kernel‑smoothed posterior, while MoE/FFN approximates it in a relaxed network‑based way;
the residual‑RMSNorm block acts as a relaxed retraction that maintains spherical geometry.

To verify the SVFlow interpretation of Transformers, experiments on pre‑trained language models are conducted using a prefix‑shuffling perturbation that progressively degrades context.
For attention layers, the relation between SVFlow‑induced metrics and performance measures is analyzed.
The results reveal that
(i) the marginal likelihood strongly correlates with predictive perplexity and calibration, serving as a reliable indicator of representation quality;
(ii) deep attention layers are far more sensitive to context disruption than shallow layers, revealing depth‑dependent specialization;
(iii) the interplay between divergence and concentrations characterizes the internal dynamics of attention,
giving rise to three distinct regimes that directly reflect calibration differences across models.
These findings directly support our framework as a lens for understanding the underlying mechanisms of Transformer.

\section{Preliminaries}

\subsection{Score-based ODE Flow}
Consider the Neural ODEs framework~\citep{chen2018neural}, where
a time-dependent diffeomorphism $\varphi_t: [0, T] \times \mathbb{R}^d \to \mathbb{R}^d$ evolves according to:
\begin{equation}
    \frac{d}{dt}\varphi_t(x) = v_t(\varphi_t(x)), \quad \varphi_0(x) \sim p_0,
\label{eq:ode}
\end{equation}
with $v_t: [0, T]\times\mathbb{R}^d \to \mathbb{R}^d$ a learnable vector field.
The log-density $\log p_t(x_t)$ along the trajectory satisfies the instantaneous change of variables formula,
with $p_t = [\varphi_t]_* p_0$ the push-forward measure.

A core insight from score-based generative modeling~\citep{song2021scorebased} is that if the vector field equals the score function of the marginal density,
i.e., $v_t(x) = \nabla_x \log p_t(x)$,
then the flow transforms samples following the steepest ascent of the log-density.

\subsection{Optimization on Spherical Manifolds}
\label{sec:spherical_optim}
Transformers inherently operate on normalized representations due to LayerNorm~\citep{ba2016layernormalization} or RMSNorm~\citep{zhang2019rms} operations,
motivating optimization on the unit sphere $\mathbb{S}^{d-1} = \{x \in \mathbb{R}^d \mid \lVert x \rVert = 1\}$.

For $x\in\mathbb{S}^{d-1}$, the tangent space $\mathcal{T}_x\mathbb{S}^{d-1} = \{v \in \mathbb{R}^d \mid x^\top v = 0\}$ contains all vectors orthogonal to $x$.
Given a tangent vector $v \in \mathcal{T}_x\mathbb{S}^{d-1}$, the exponential map $\exp_x: \mathcal{T}_x\mathbb{S}^{d-1} \to \mathbb{S}^{d-1}$ provides the canonical geodesic step:
\begin{equation}
\exp_x(v) = \cos(\lVert v \rVert) x + \sin(\lVert v \rVert) \frac{v}{\lVert v \rVert}, \quad v \neq 0.
\label{eq:exponential_map}
\end{equation}

While exact, the trigonometric computations can be inefficient. A computationally efficient approximation, known as retraction, is widely used in manifold optimization:
\begin{equation}
R_x(v) = \frac{x + v}{\lVert x + v \rVert}, \quad v\in\mathcal{T}_x\mathbb{S}^{d-1},
\label{eq:retraction}
\end{equation}
which satisfies $R_x(v) = \exp_x(v) + \mathcal{O}(\lVert v\rVert^2)$.

\subsection{Transformer Building Blocks}
Transformers~\citep{vaswani2017attention} are composed of Multi-Head Attention (MHA), Feed-Forward Networks (FFN),
residual connections and normalization layers.

\textbf{Multi-Head Attention.}
Given a query vector $x_q \in \mathbb{R}^d$ and key-value pairs $\{(k_z, v_z)\}_{z\in\mathcal{Z}}$ with finite index set $\mathcal{Z}$,
the $\ell$-th Transformer layer updates $x_q$ via residual connection $ x_q^{\ell+1} = x_q^{\ell} + f_\ell(x_q^{\ell})$. 
For attention layers, $f_\ell = f_{\text{Attn}}$ is defined as:
\begin{equation}
f_{\text{Attn}}(x_q) = \sum_{h=1}^{H} W_{o,h} \left( \sum_{z\in\mathcal{Z}} \sigma_{k_z,h}(x_q) \cdot W_{v,h}^\top v_z \right)
= \sum_{h=1}^{H} \sum_{z\in\mathcal{Z}} \sigma_{k_z,h}(x_q) \cdot \left(W_{o,h} W_{v,h}^\top\right) v_z.
\label{eq:mha}
\end{equation}
Here $H$ is the number of attention heads, $W_{o,h}, W_{v,h} \in \mathbb{R}^{d \times d'}$ are projection matrices,
and the attention weight is:
\begin{equation}
    \sigma_{k_z,h}(x_q) = \frac{\exp\left( k_z^\top \left(W_{k,h} W_{q,h}^\top\right) x_q \right)}{\sum_{z^\prime\in\mathcal{Z}} \exp\left( k_{z^\prime}^\top \left(W_{k,h} W_{q,h}^\top\right) x_q \right)},
\label{eq:attn_softmax}
\end{equation}
with $W_{k,h}, W_{q,h} \in \mathbb{R}^{d \times d'}$ key and query projections.

\textbf{Feed-Forward Networks and Mixture of Experts.}
The standard FFN is a special case of Mixture of Experts (MoE)~\citep{shazeer2017,jacobs1991adaptive} with a single expert.
General MoE layers compute:
\begin{equation}
f_{\text{MoE}}(x) = \sum_{i=1}^{E} g_i(x) \cdot e_i(x),
\label{eq:moe}
\end{equation}
where $E$ is the number of experts, $e_i(x)$ is the $i$-th expert network,
and $g_i(x)$ are gating weights satisfying $g_i(x) \geq 0$ and $\sum_i g_i(x) = 1$.

\textbf{Spherical Normalization via RMSNorm.}
Modern Transformers employ various normalization schemes to stabilize training.
Of particular relevance to our geometric framework is RMSNorm~\citep{zhang2019rms}:
\begin{equation}
\label{eq:rmsnorm}
\text{RMSNorm}(x) = \sqrt{d}\cdot \frac{x}{\lVert x \rVert},
\end{equation}
which projects $x$ onto the sphere $\mathbb{S}^{d-1}(\sqrt{d})$ of radius $\sqrt{d}$.
Combined with residual connection, the layer update becomes:
\begin{align}
\label{eq:euler_step}
  x^{\ell+1}=\text{RMSNorm}\left(x^\ell + f_\ell(x^\ell)\right).
\end{align}

\section{Score-based Variational Flow}
\label{sec:svflow_framework}

\subsection{Definition of SVFlow}
\label{sec:SVFlow}
Let $x\in\mathbb{R}^d$ be a continuous random variable, and $z\in\mathcal{Z}$ a discrete latent variable
such that the marginal distribution of $x$ follows the mixture property $p(x)=\mathbb{E}_{p(z)}\left[p(x\vert z)\right]$.
Given a $\theta$-parameterized conditional likelihood $p(x\vert z;\theta)$ and a $\phi$-parameterized variational posterior $q(z\vert x;\phi)$,
we define:
\begin{definition}[SVFlow]
A deterministic ODE flow is a \textbf{Score-based Variational Flow (SVFlow)} if its time-dependent vector field is
the variational posterior-weighted average of conditional score fuctions.
Formally, for $t\in [0, T]$, the flow is defined by:
\begin{equation}
  \frac{dx_t}{dt} = v_t(x_t)= \mathbb{E}_{q_t(z\vert x; \phi_t)}\left[\nabla_{x}{\log p_t(x\vert z; \theta_t)}\right],
\label{eq:variational_flow}
\end{equation}
where $q_t(z\vert x; \phi_t)$ and $p_t(x\vert z;\theta_t)$ are time-evolving variational posterior and
conditional likelihood families respectively, with parameters $\phi_t$ and $\theta_t$ varying continuously with $t$.
The initial condition $x_0$ is drawn from a given initial measure $p_0$.
\end{definition}
Intuitively, SVFlow is a specialized neural ODE~\citep{chen2018neural} with interpretable structural constraints.
The evolution is guided by two complementary mechanisms:
(i) \textbf{variational posterior weighting}:
 the expectation over $q_t(z\vert x;\phi_t)$ assigns higher weights to latent variables $z$ that are more plausible given the current state,
 acting as a dynamic filtering mechanism;
(ii) \textbf{conditional likelihood ascent}:
 the conditional score drives $x_t$ toward regions of high conditional density for each component.

\subsection{Properties of SVFlow}
The marginal log-density $\log p_t(x)$ along SVFlow's trajectory admits a tight variational lower bound:
\begin{proposition}[SVFlow Evidence Lower Bound]
\label{prop:elbo_bound}
For all $t\in [0, T]$, the marginal log-density along the trajectory of SVFlow satisfies:
\begin{equation}
\log p_t(x) = \mathcal{L}_t(x) + \mathrm{KL}(q_t(z\vert x)\Vert p_t(z\vert x))
            \geq \mathcal{L}_t(x)
\end{equation}
where $\mathrm{KL} \geq 0$ is the Kullback-Leibler divergence,
and $\mathcal{L}_t$ is the \textbf{instantaneous Evidence Lower Bound (ELBO)} defined as:
\begin{equation}
\mathcal{L}_{t}(x)\triangleq\mathbb{E}_{q_t(z\vert x)}\left[\log p_t(x\vert z)\right]-\mathrm{KL}\left(q_t(z\vert x)\Vert p_t(z)\right).
\label{eq:elbo}
\end{equation}
Here $p_t(z)$ denotes the prior distribution over $z$, assumed uniform and time-invariant for simplicity.
Equality holds if and only if $q_t(z\vert x) = p_t(z\vert x)$.
\end{proposition}
This variational bound connects SVFlow to classic variational inference~\citep{blei2017variational}.

\begin{theorem}[Gradient Decomposition]
\label{thm:grad_decomp}
The SVFlow vector field $v_t(x)$ and the instantaneous ELBO $\mathcal{L}_t(x)$ satisfy:
\begin{equation}
\label{eq:grad_decomp}
  \nabla_x\mathcal{L}_t(x) = v_t(x) + \underbrace{\mathbb{E}_{q_t(z\vert x)}\left[\log\frac{p_t(z\vert x)}{q_t(z\vert x)}\nabla_x\log q_t(z\vert x)\right]}_{\epsilon_t(x)},
\end{equation}
where $\epsilon_t(x)$ quantifies the variational approximation error and vanishes if and only if $q_t(z|x) = p_t(z|x)$.
\end{theorem}

Thus, $v_t(x)$ is the conditional score component of $\nabla_x \mathcal{L}_t(x)$,
while $\epsilon_t(x)$ biases the flow away from the steepest ELBO ascent when the posterior is misaligned.
A direct corollary characterizes the ideal regime without this bias:
\begin{corollary}[Ideal SVFlow Conditions]
\label{corollary:ideal}
If $q_t(z\vert x) = p_t(z\vert x)$ for all $t$, then $\epsilon_t(x)=0$, and
\begin{equation}
    v_t(x) = \mathbb{E}_{p_t(z\vert x)}\left[\nabla_x\log p_t(x\vert z)\right]
           =\nabla_x\mathcal{L}_t(x) = \nabla_x\log p_t(x).
\end{equation}
\end{corollary}
In this ideal regime, the SVFlow trajectory follows the steepest ascent of the true marginal log-probability $\log p_t(x)$.

\subsection{Training Objective of SVFlow}
Building upon the theoretical properties established above,
we now derive a training objective that unifies variational consistency regularization with semantic alignment supervision.

\textbf{Variational consistency.}
From Proposition~\ref{prop:elbo_bound}, the gap between $\log p_t(x)$ and $\mathcal{L}_t(x)$ is exactly $\mathrm{KL}\left(q_t(z\vert x)\Vert p_t(z\vert x)\right)$.
Minimizing this divergence aligns $q_t(z\vert x)$ with $p_t(z\vert x)$, thereby
eliminating the approximation error $\epsilon_t(x)$ and steering the SVFlow vector field toward the true score.
We define the variational consistency objective over the SVFlow trajectory:
\begin{equation}
\label{eq:var_consistency}
    \mathcal{J}_\text{var}(\theta,\phi) = \mathbb{E}_{t\sim\mathcal{U}(0,T),x_0\sim p_0}\left[\lambda(t)\cdot\mathrm{KL}\left(q_t(z\vert x_t)\Vert p_t(z\vert x_t)\right)\right].
\end{equation}
Here $\lambda: [0,T]\to \mathbb{R}_+$ is a time-dependent weighting function satisfying $\mathbb{E}_{t\sim\mathcal{U}(0, T)}[\lambda(t)] = 1$,
used for focusing regularization on certain parts of the flow.
Optimizing $\mathcal{J}_\text{var}$ alone, however, tends to collapse the posterior to a single component, losing input‑dependent filtering.

\textbf{Semantic alignment.} To inject task‑specific information, we assume access to paired data $(x_0, y)\sim q_\text{data}(x, y)$.
The \emph{semantic alignment objective} forces the terminal representation $x_T$ to capture the relevant aspects of $y$ with negative log-likelihood loss:
\begin{equation}
\label{eq:align_obj}
\mathcal{J}_\text{align}(\theta, \phi)
    = \mathbb{E}_{x_0\sim q_\text{data}(x)}\left[\mathrm{KL}\left(q_\text{data}(y\vert x_0)\Vert p_T(y\vert x_T)\right)\right]
    = \mathbb{E}_{x_0, y}\left[-\log p_T(y\vert x_T)\right] + \text{const.}
\end{equation}
where $p_T(y\vert x_T)$ is a task‑specific prediction model, e.g., a softmax classifier for discrete labels.
This objective drives the flow to produce linearly separable representations for discriminative tasks.

\begin{figure*}[t]
  \vspace{-3mm}
  \centering
  \begin{minipage}[c]{0.23\textwidth}
  \centering
  \subcaptionbox{Data\label{fig:toys_data}}[0.6\textwidth]{
      \setlength\fboxsep{0pt}
      \fbox{\includegraphics[width=\linewidth]{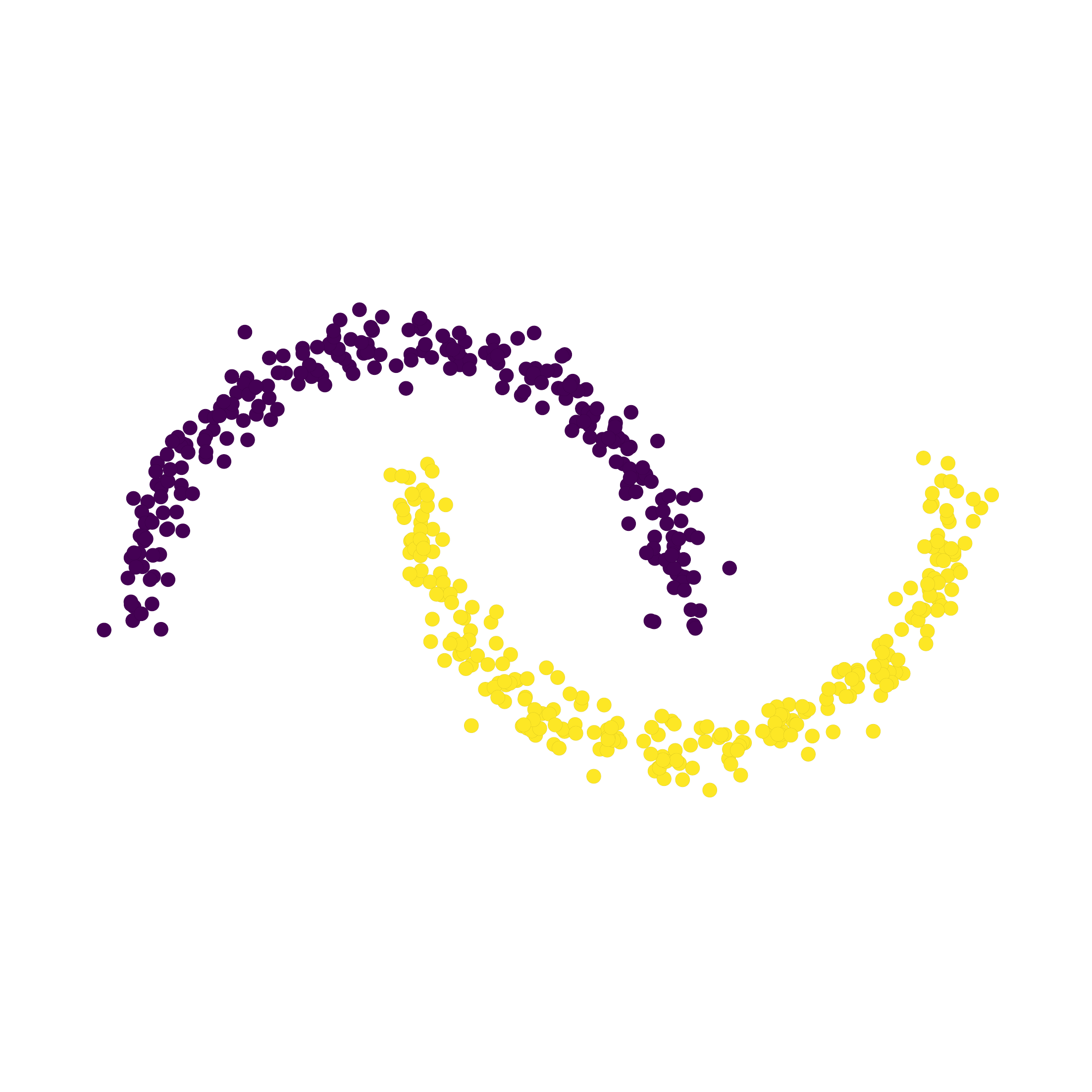}}
  }
  \vspace{2pt}
  \subcaptionbox{$\mathcal{J}_\text{var}$ only\label{fig:var_only}}[0.6\textwidth]{
      \setlength\fboxsep{0pt}
      \fbox{\includegraphics[width=\linewidth]{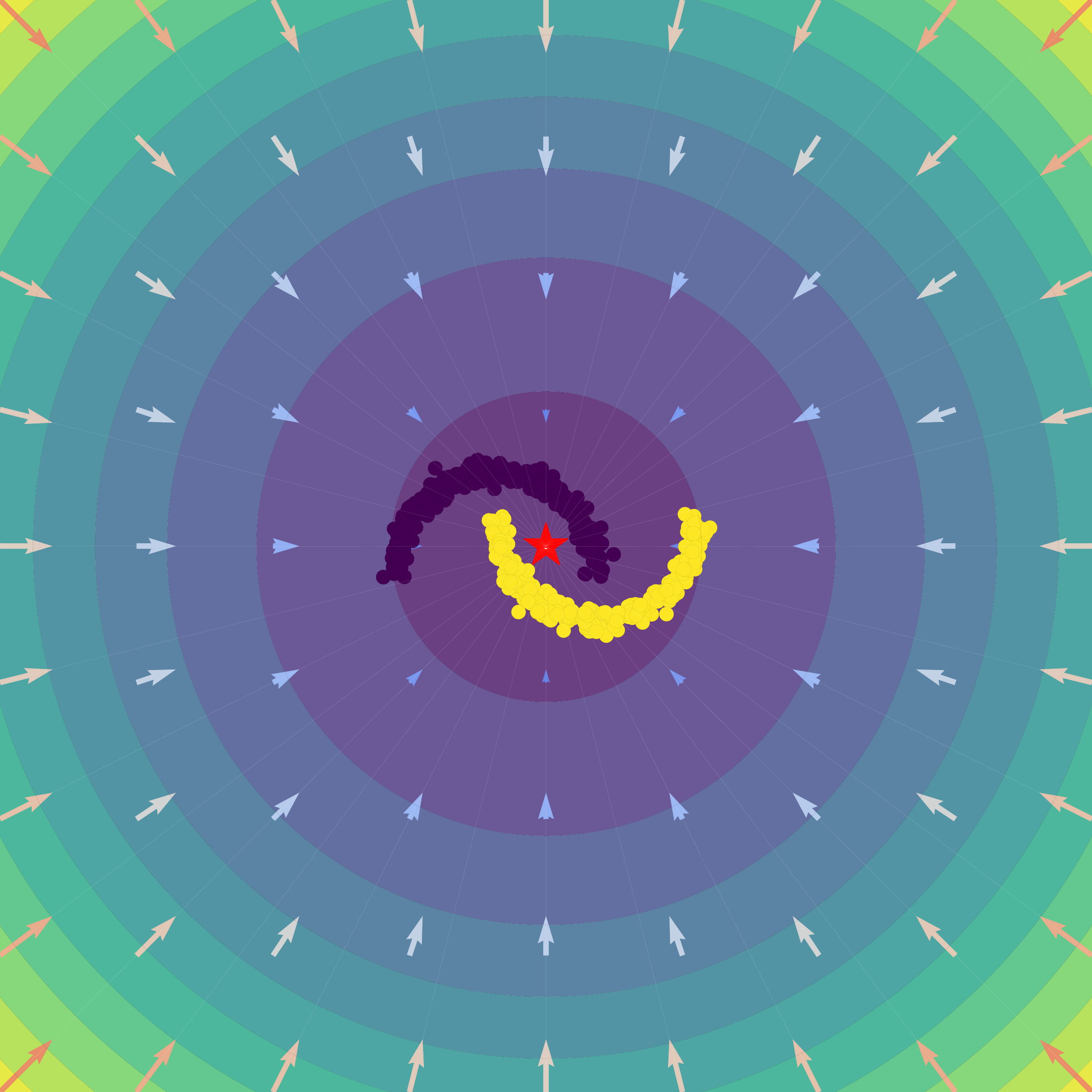}}
  }
  \end{minipage}
  \begin{minipage}[c]{0.76\textwidth}
  \centering
  \subcaptionbox{Trajectories of hybrid training under different regularization strengths\label{fig:hybrid_traj}}[0.95\linewidth]{
    \setlength{\tabcolsep}{1pt}
    \setlength\fboxsep{0pt}
    \begin{tabular}{r c c c c c}
       & $t=0.2$ & $t=0.4$ & $t=0.6$ & $t=0.8$ & $t=1.0$ \\
      \raisebox{0.2\height}{\rotatebox{90}{$\beta=0.0$}} &
        \fbox{\includegraphics[width=0.15\textwidth]{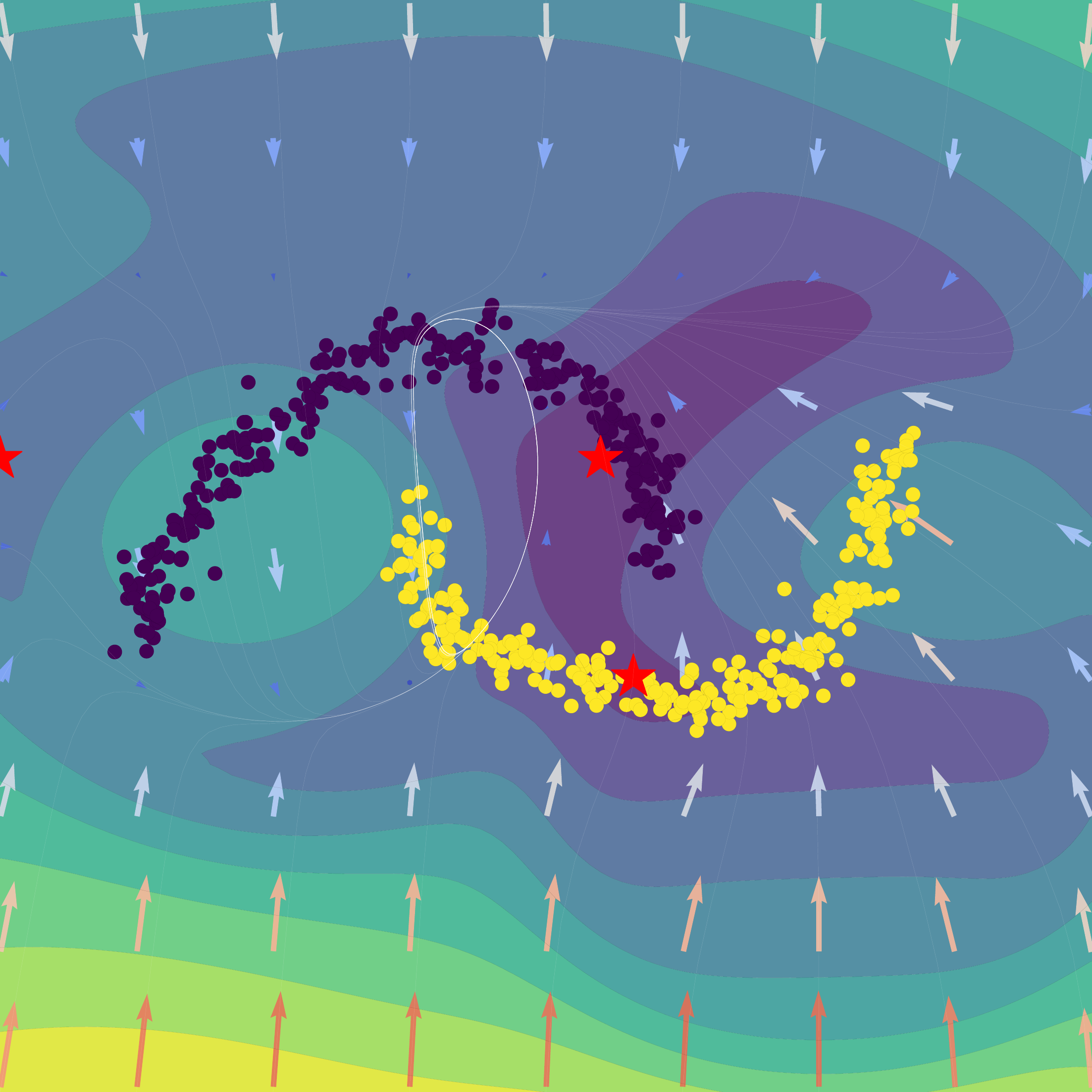}} &
        \fbox{\includegraphics[width=0.15\textwidth]{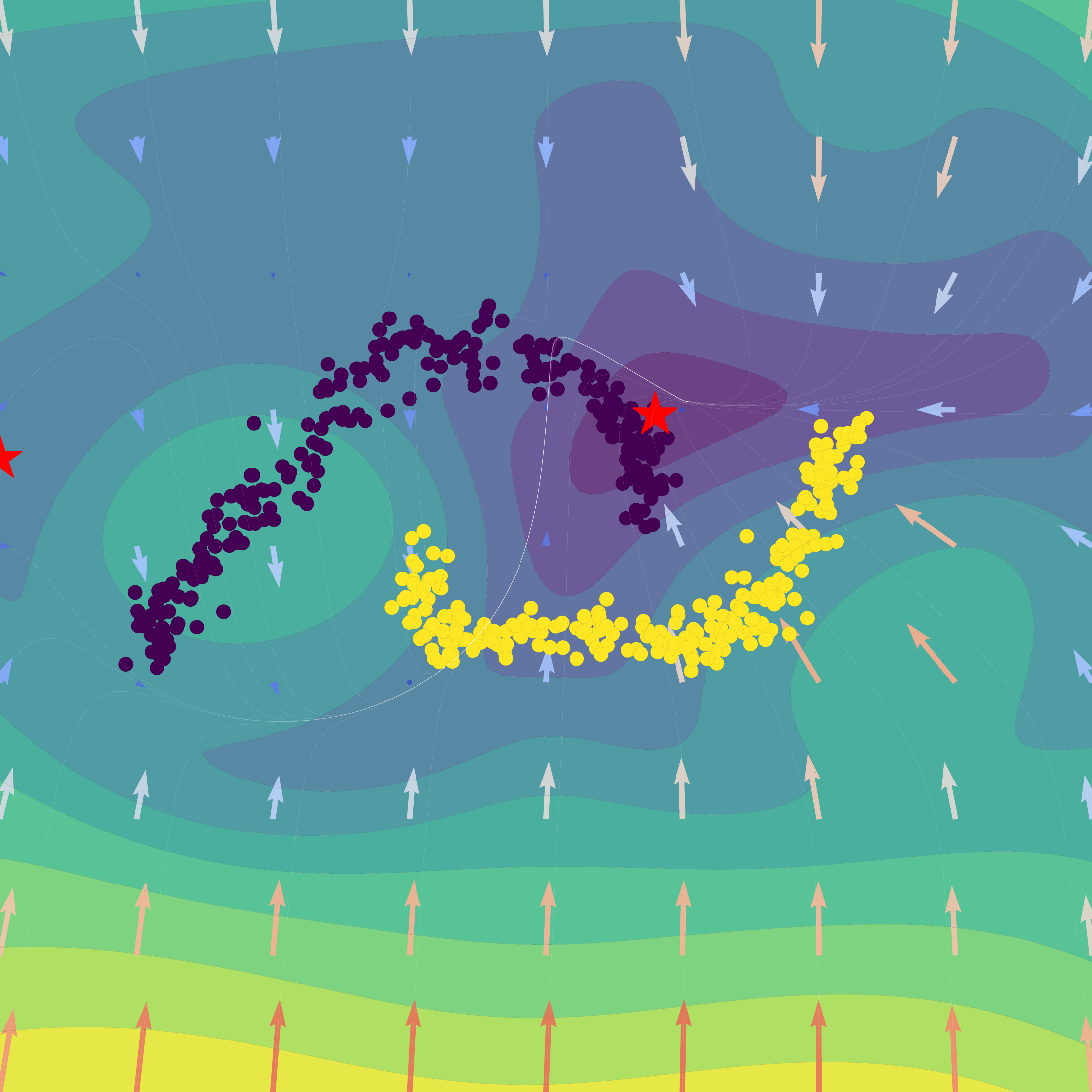}} &
        \fbox{\includegraphics[width=0.15\textwidth]{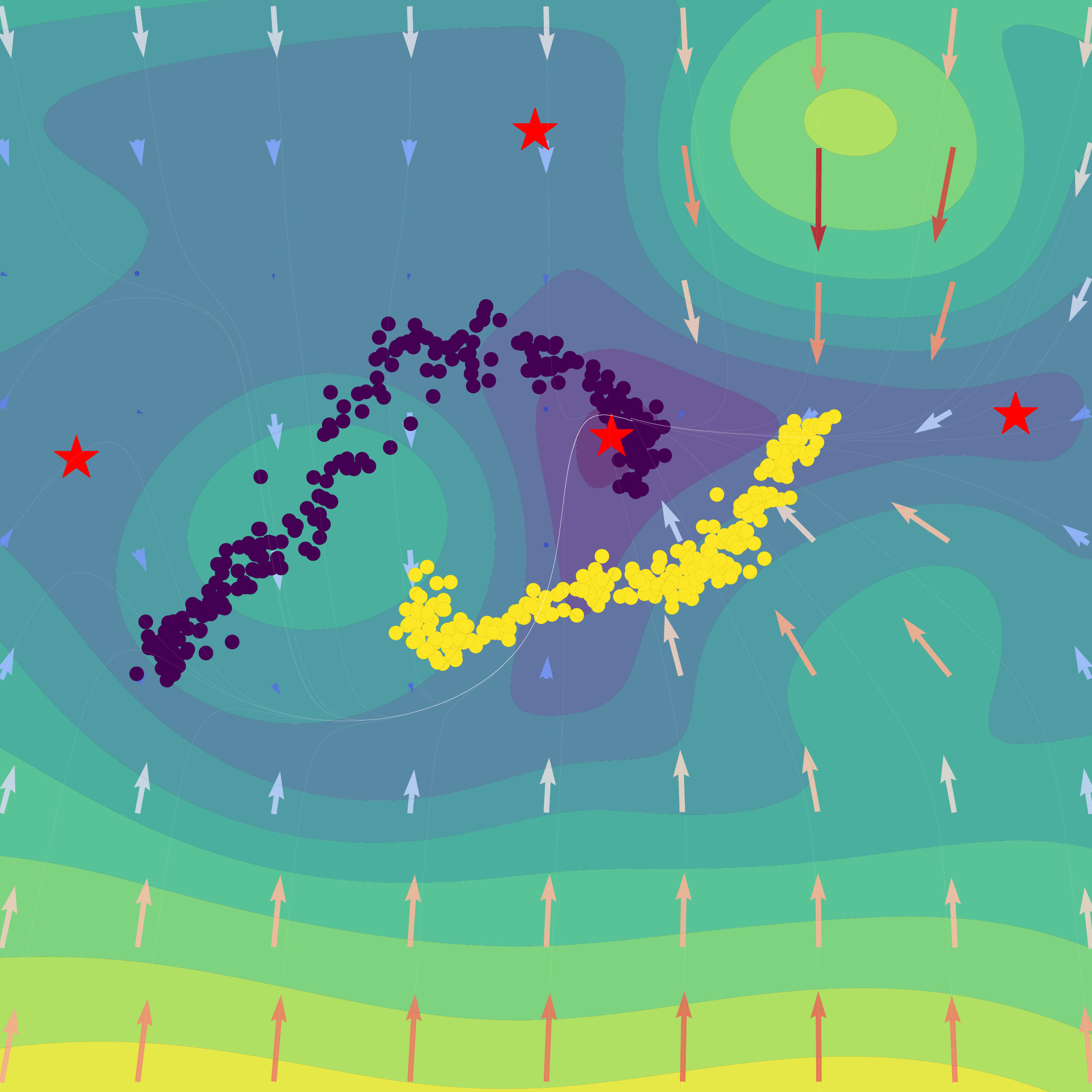}} &
        \fbox{\includegraphics[width=0.15\textwidth]{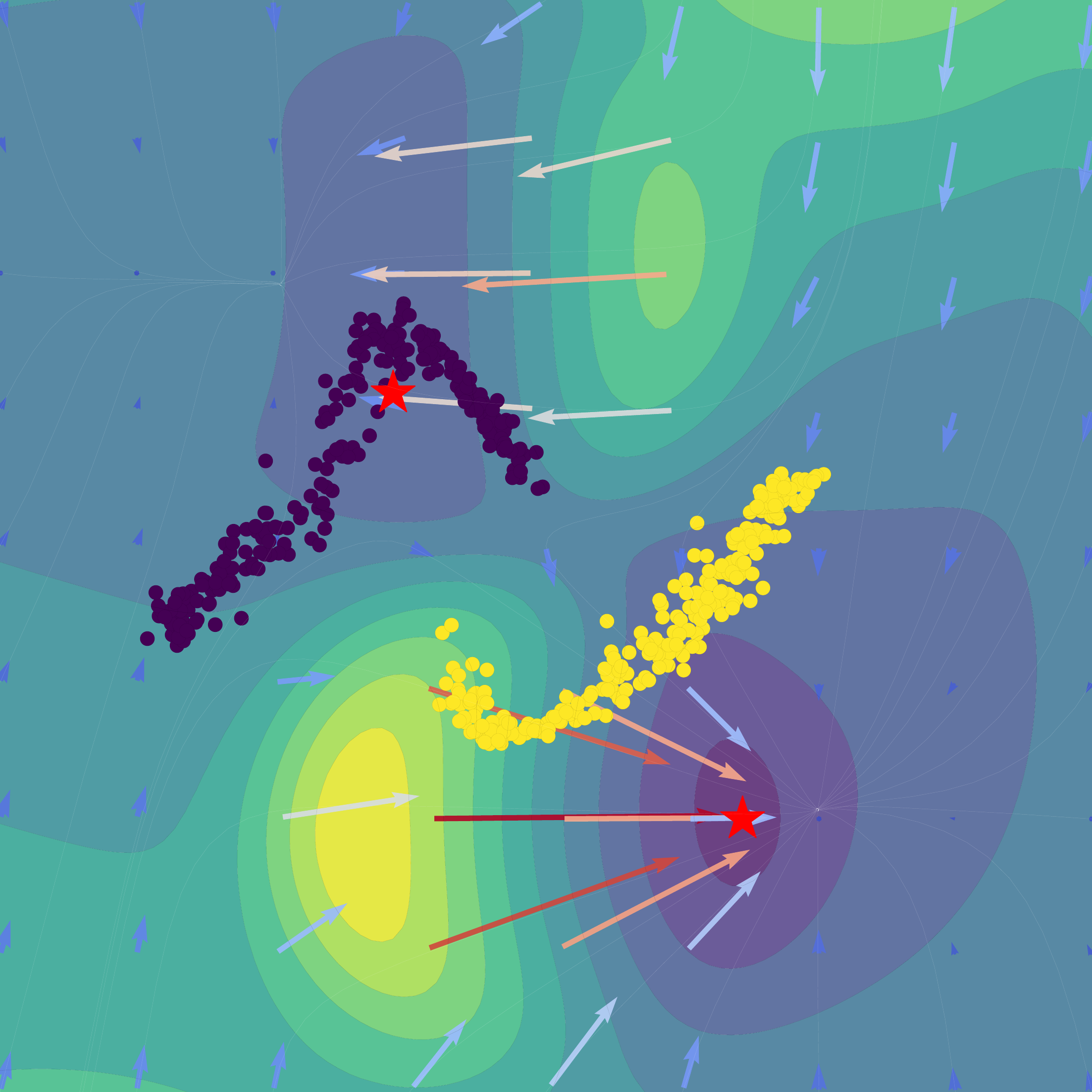}} &
        \fbox{\includegraphics[width=0.15\textwidth]{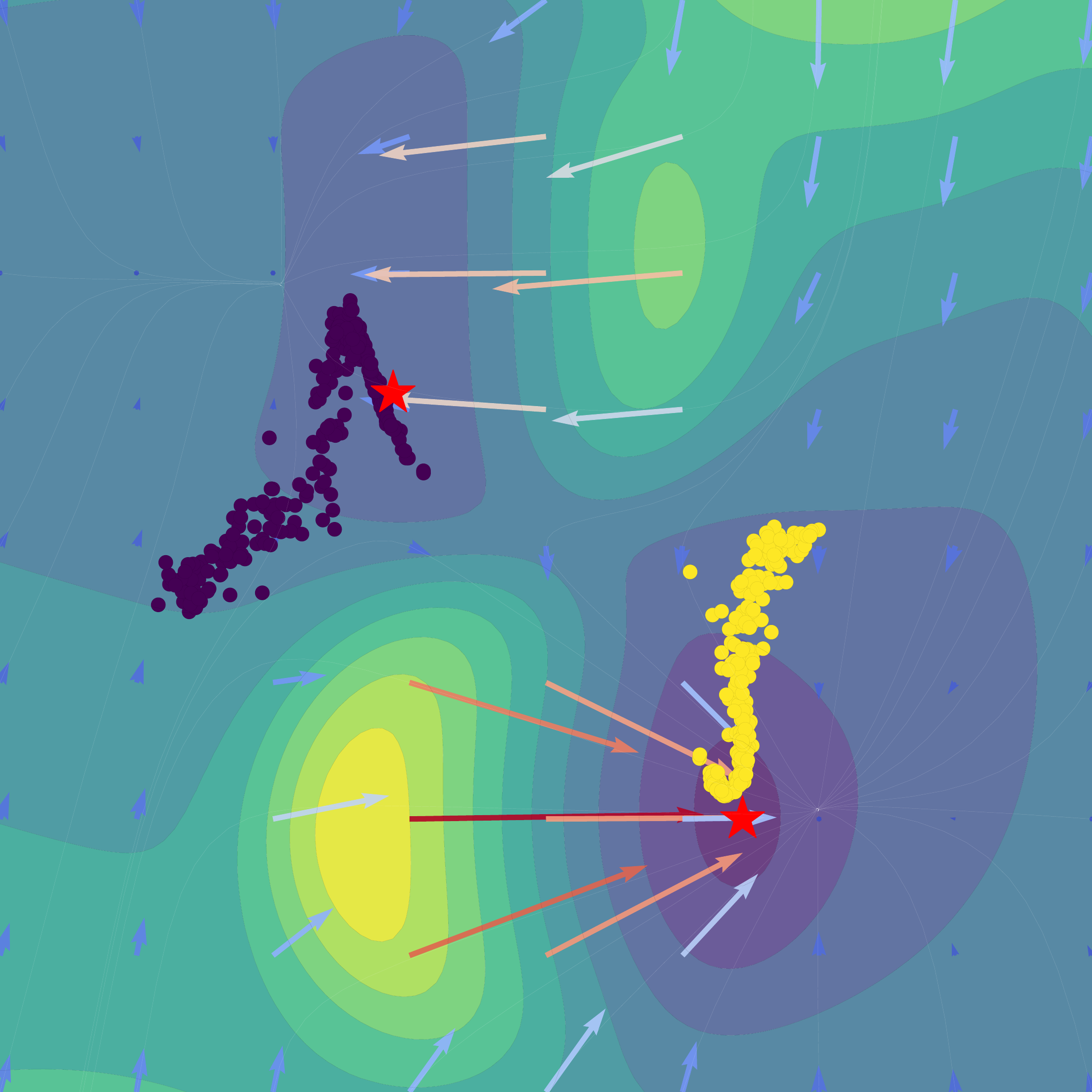}}  \\
      \raisebox{0.2\height}{\rotatebox{90}{$\beta=0.1$}} &
        \fbox{\includegraphics[width=0.15\textwidth]{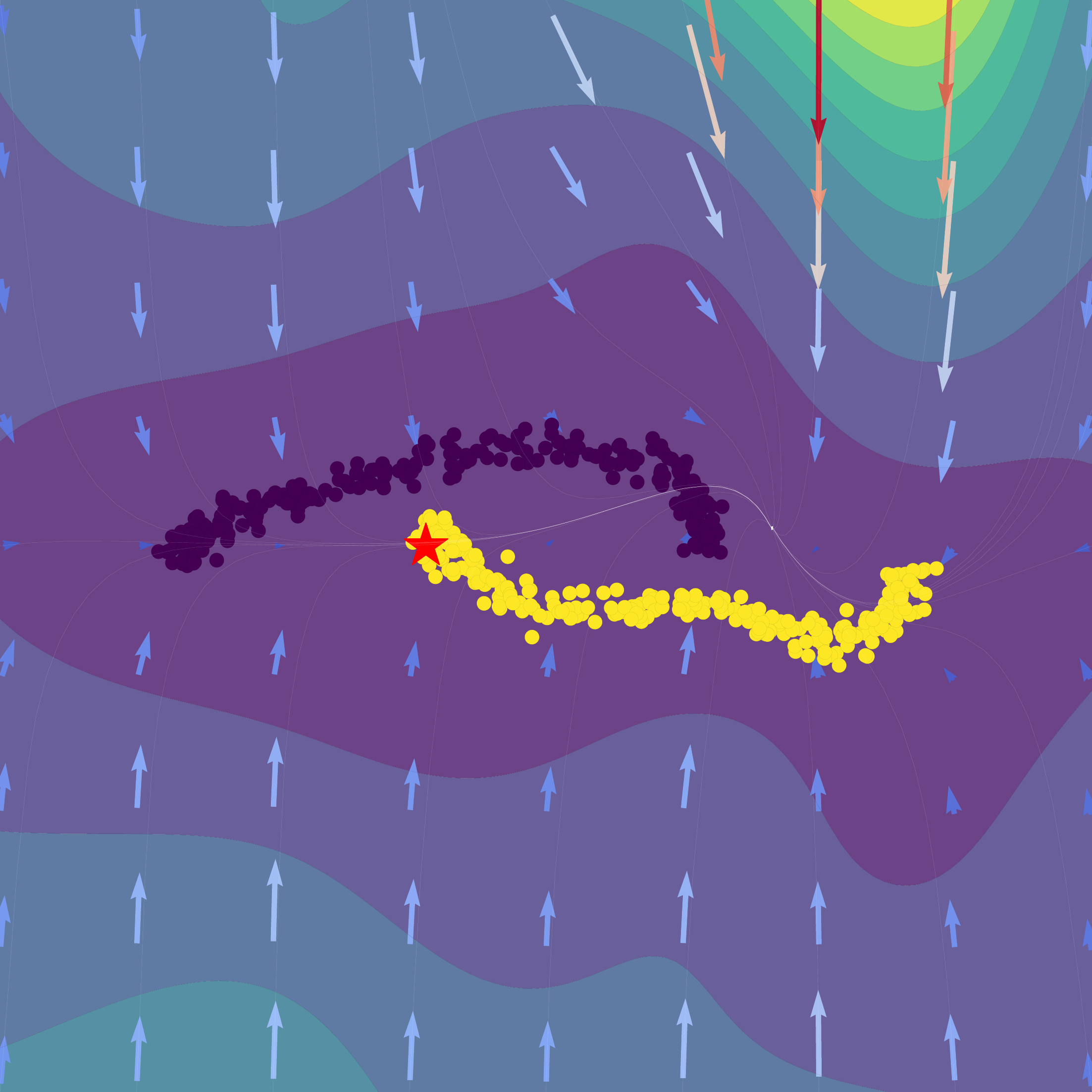}} &
        \fbox{\includegraphics[width=0.15\textwidth]{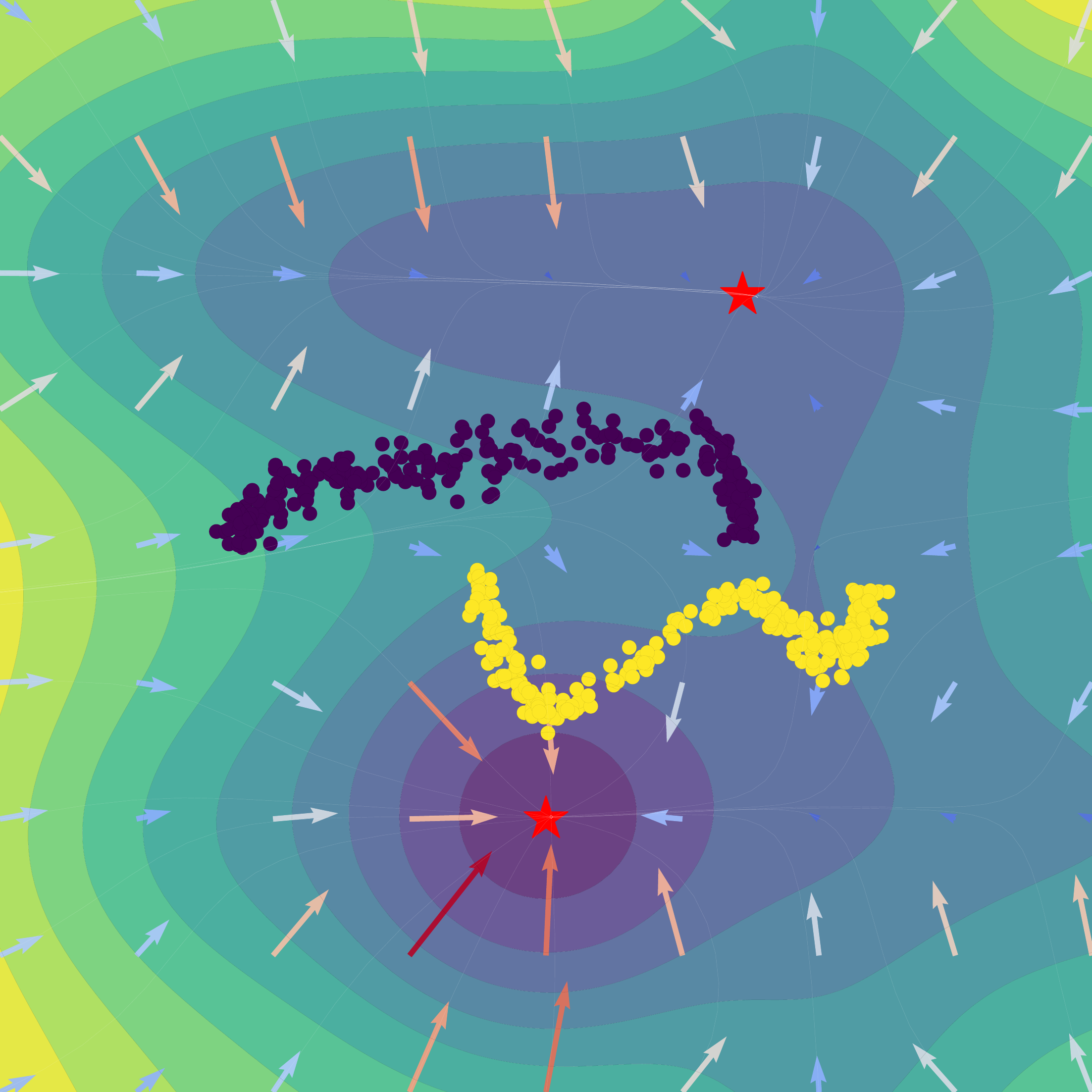}} &
        \fbox{\includegraphics[width=0.15\textwidth]{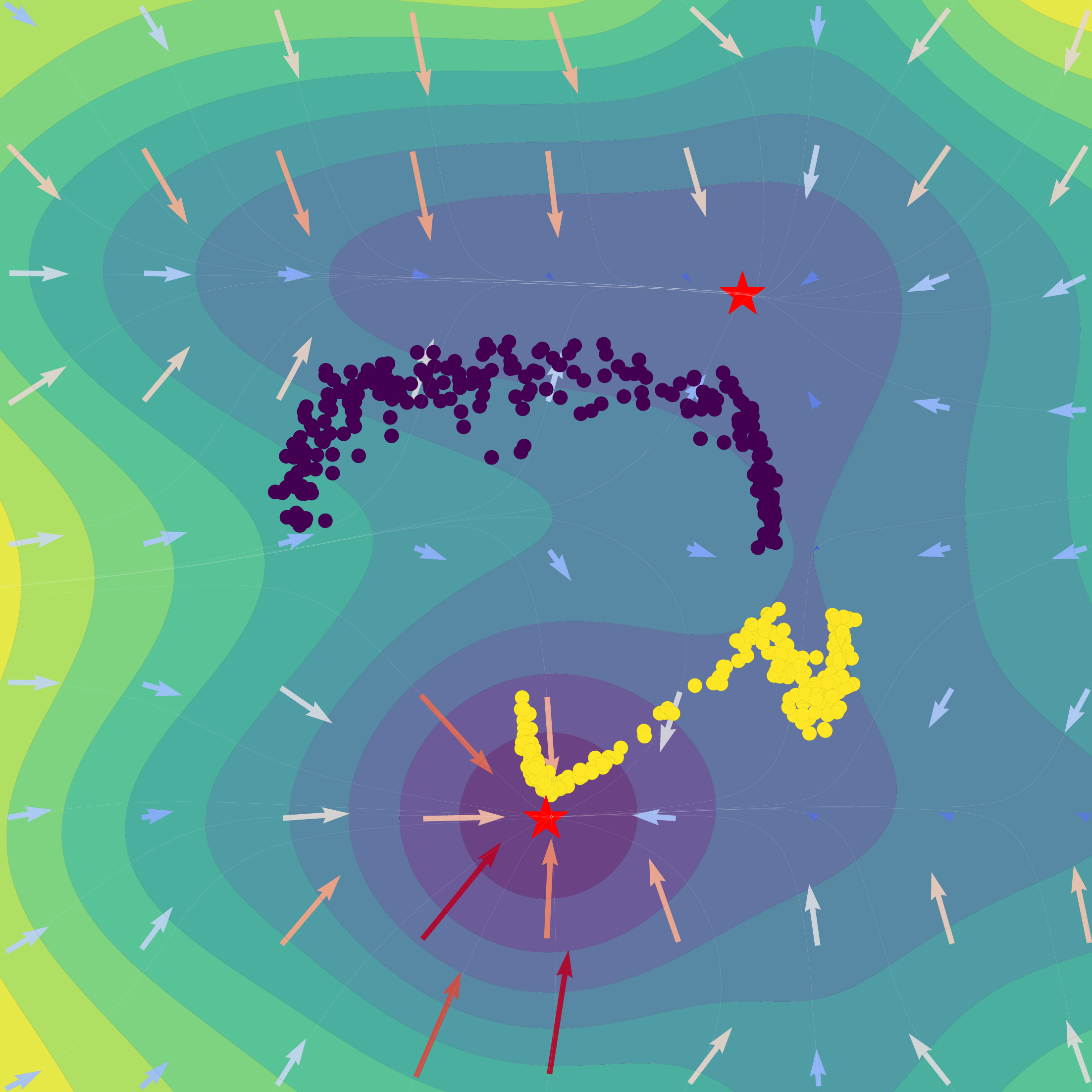}} &
        \fbox{\includegraphics[width=0.15\textwidth]{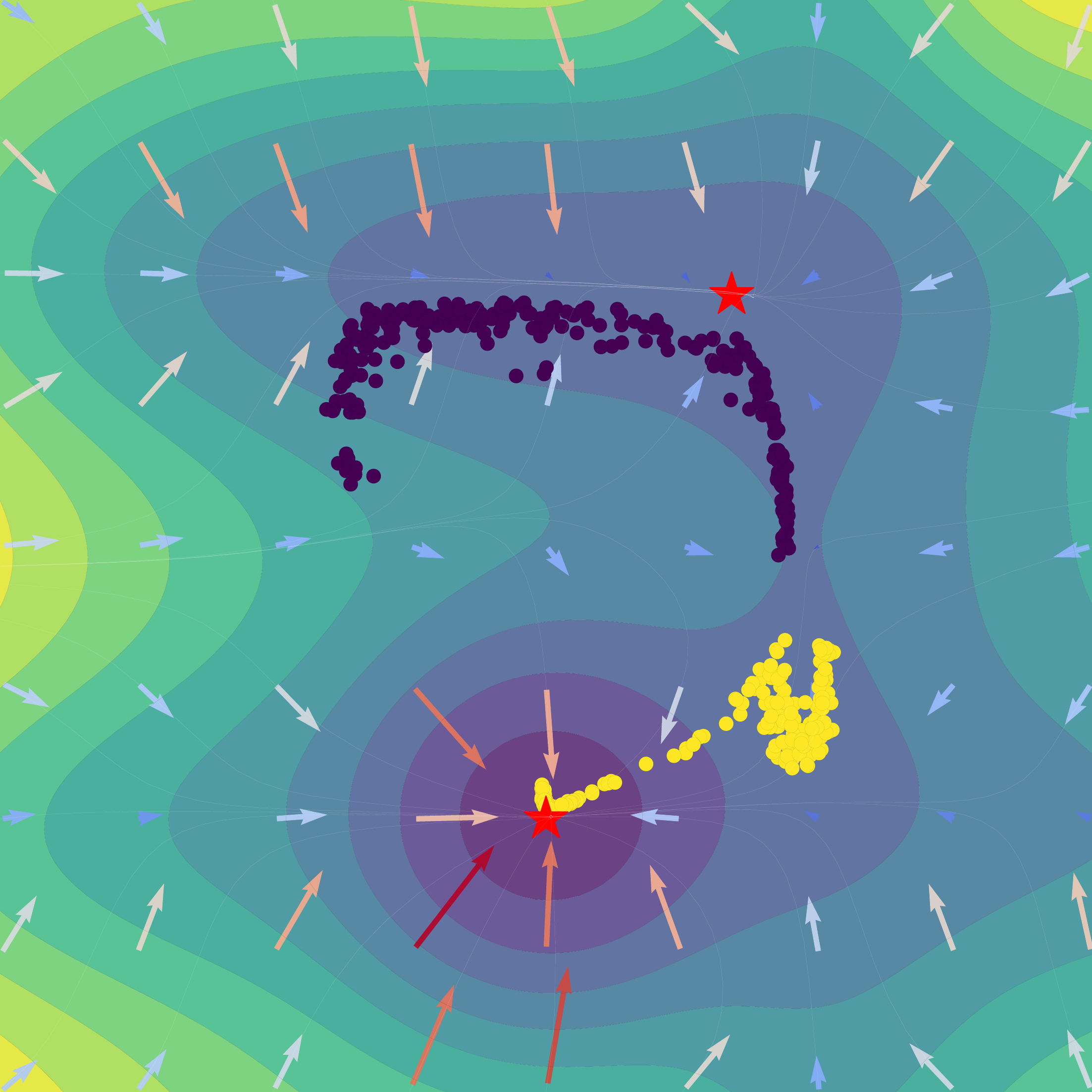}} &
        \fbox{\includegraphics[width=0.15\textwidth]{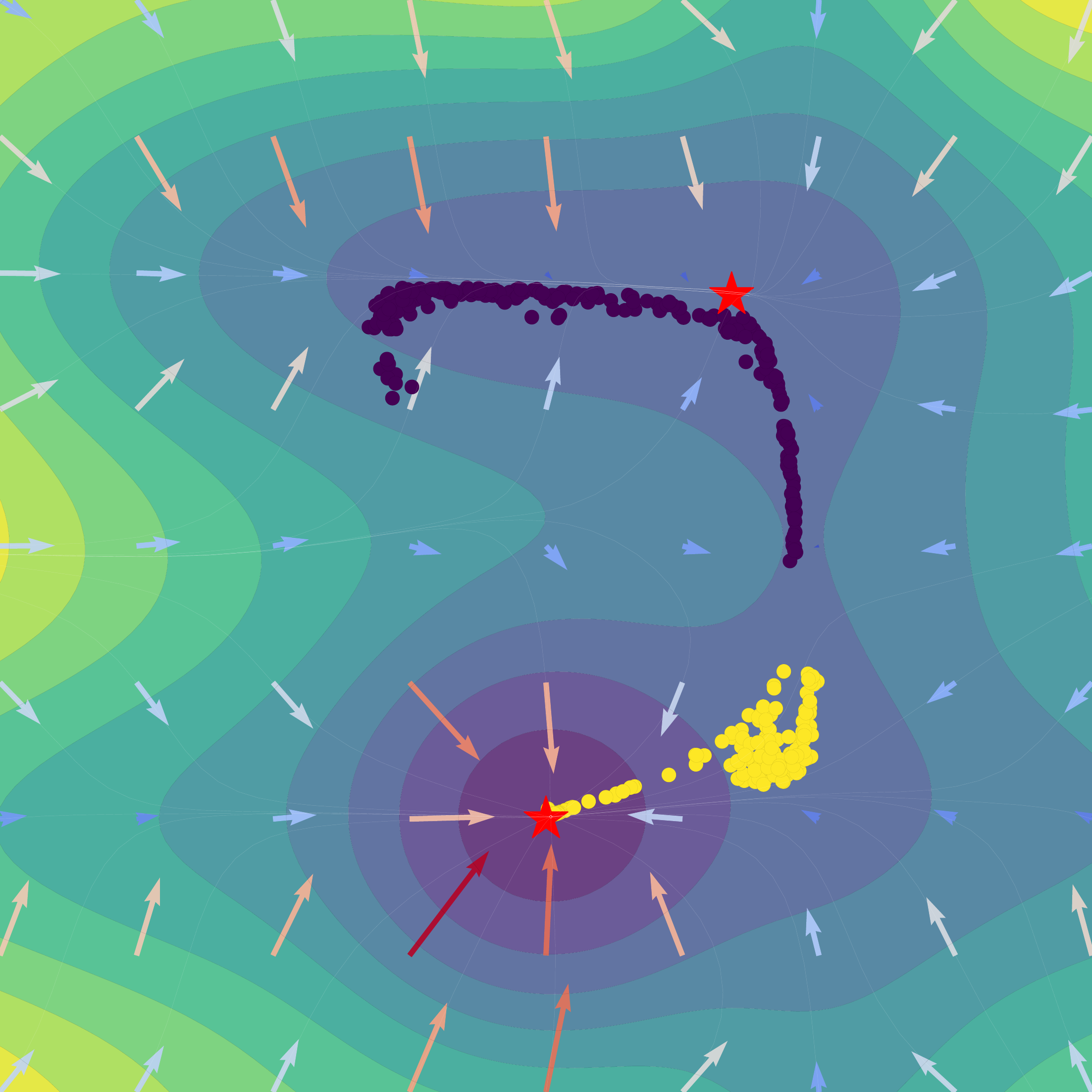}}  \\
      \raisebox{0.2\height}{\rotatebox{90}{$\beta=0.5$}} &
        \fbox{\includegraphics[width=0.15\textwidth]{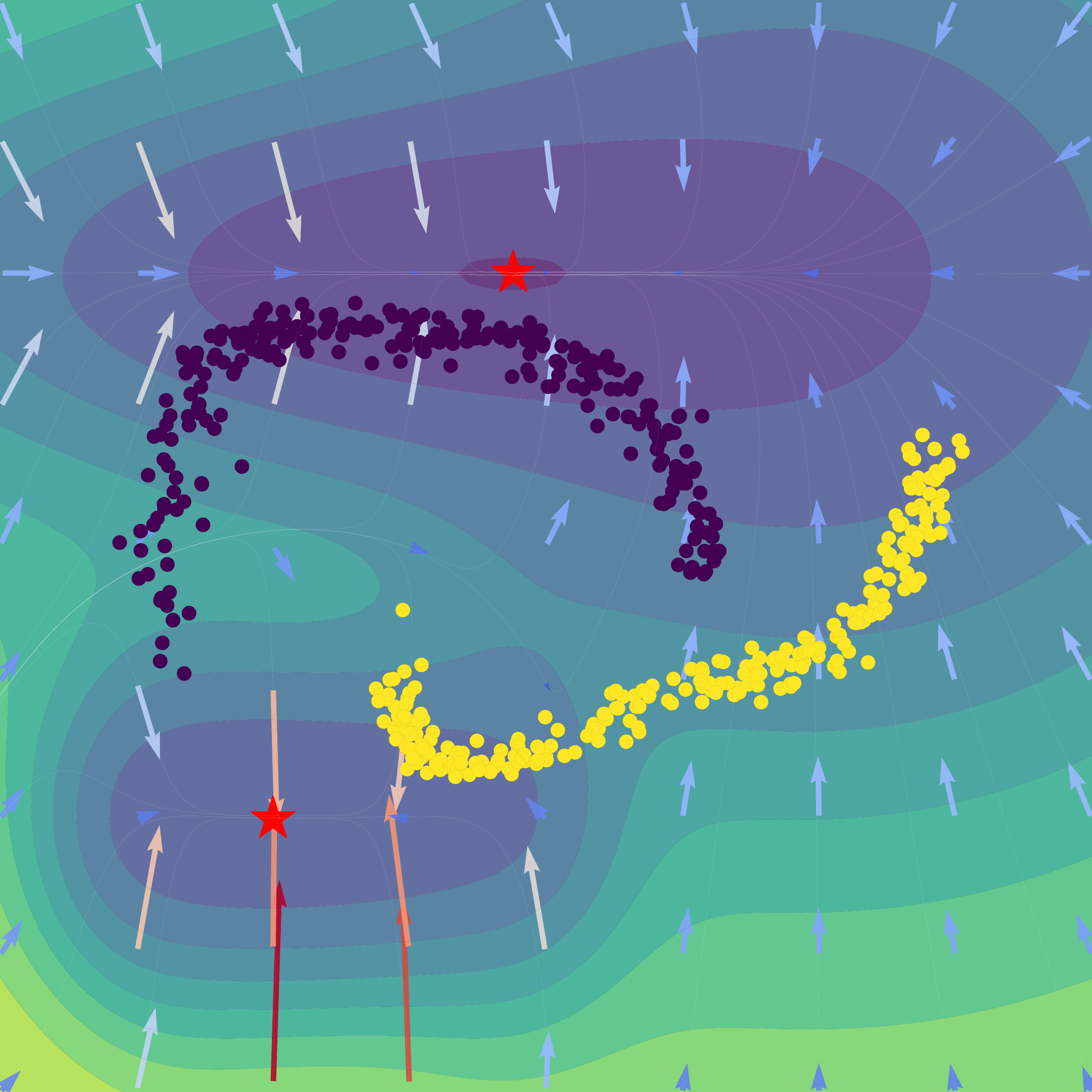}} &
        \fbox{\includegraphics[width=0.15\textwidth]{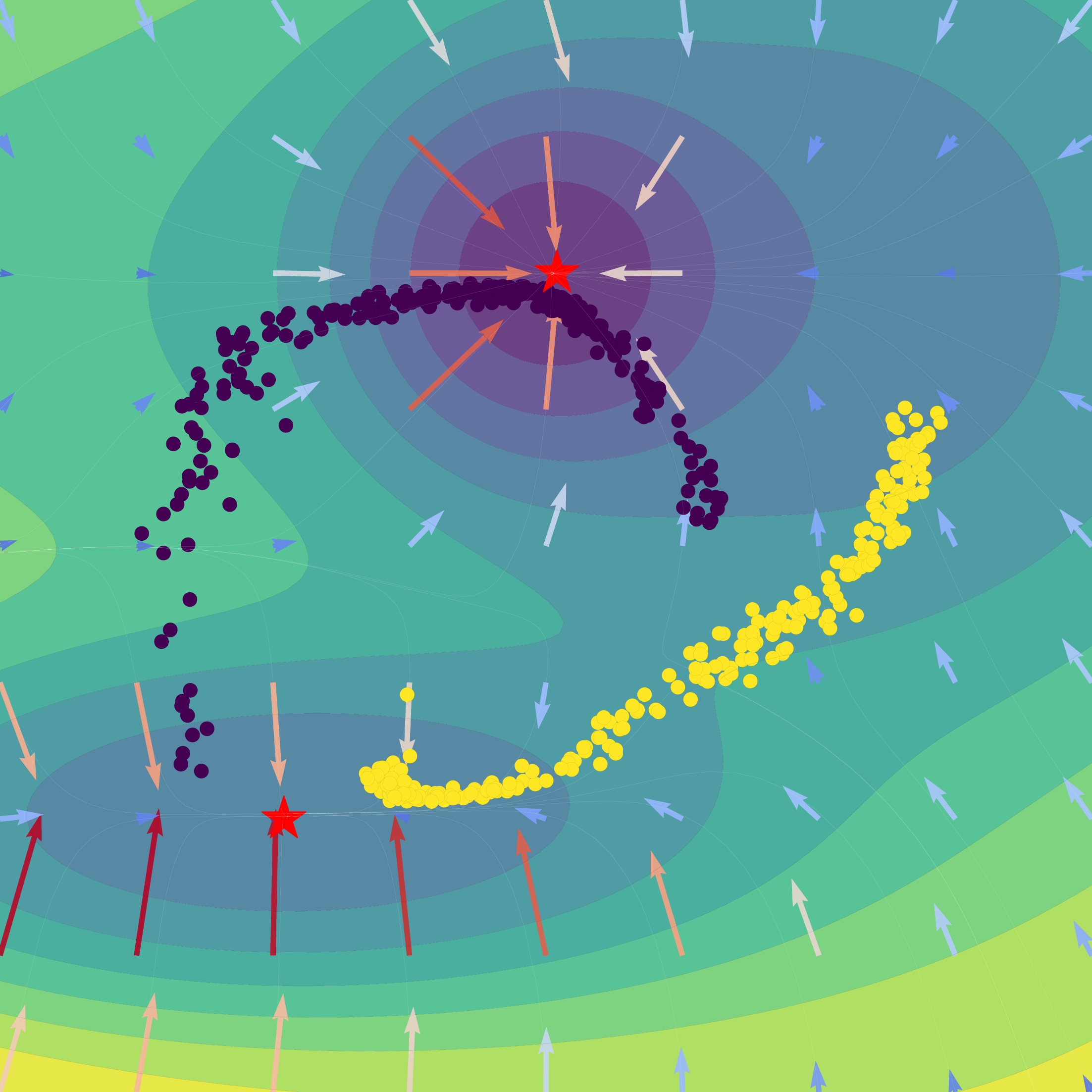}} &
        \fbox{\includegraphics[width=0.15\textwidth]{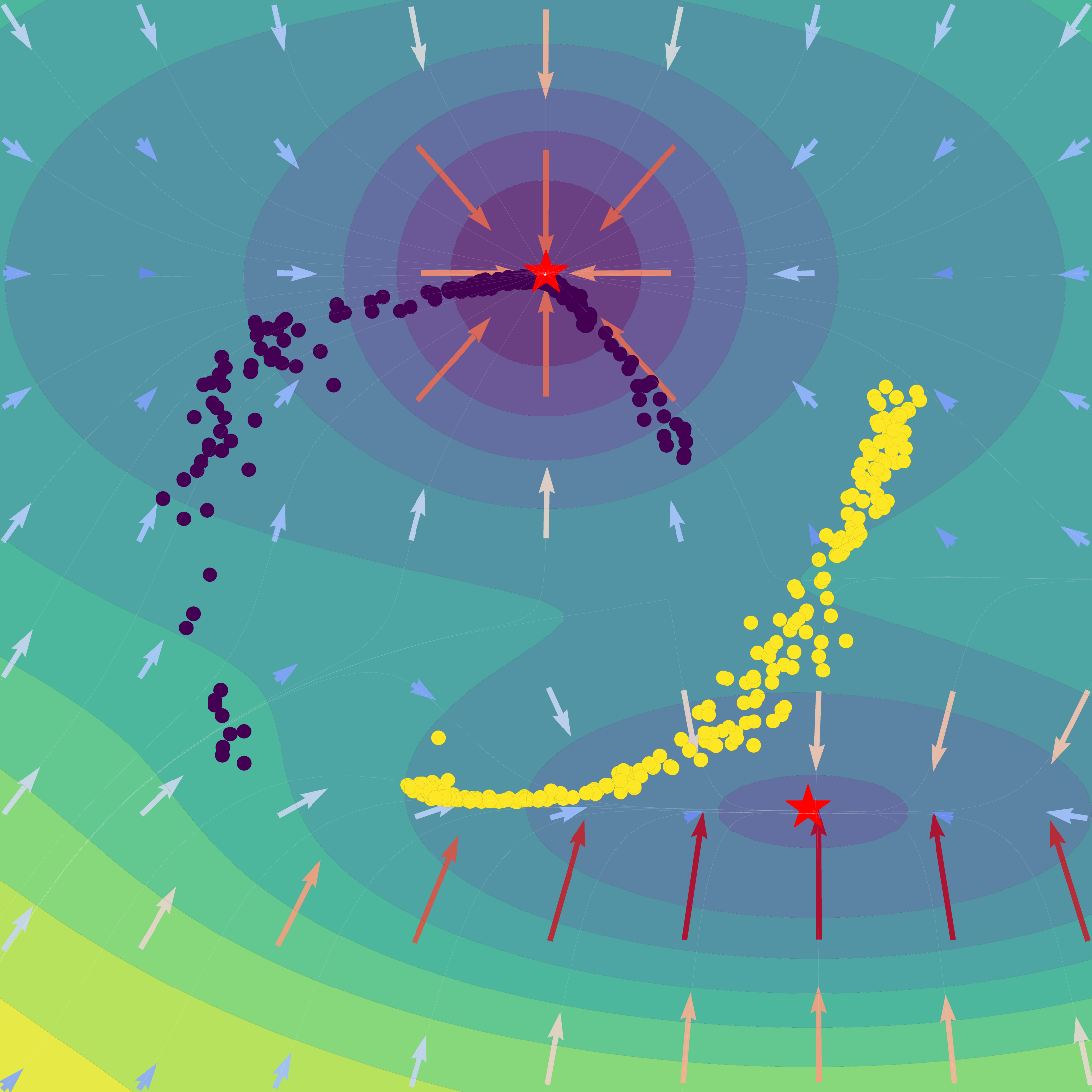}} &
        \fbox{\includegraphics[width=0.15\textwidth]{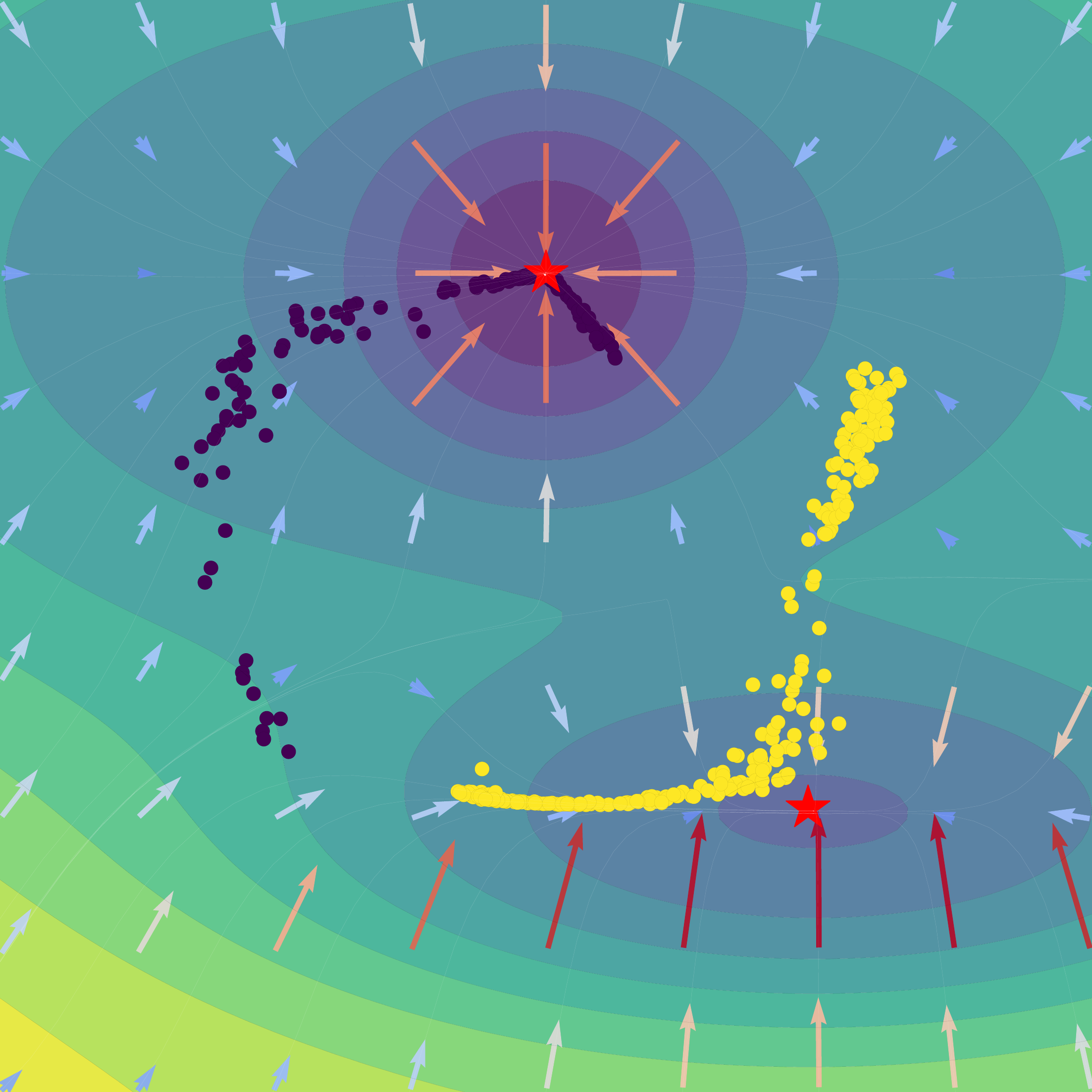}} &
        \fbox{\includegraphics[width=0.15\textwidth]{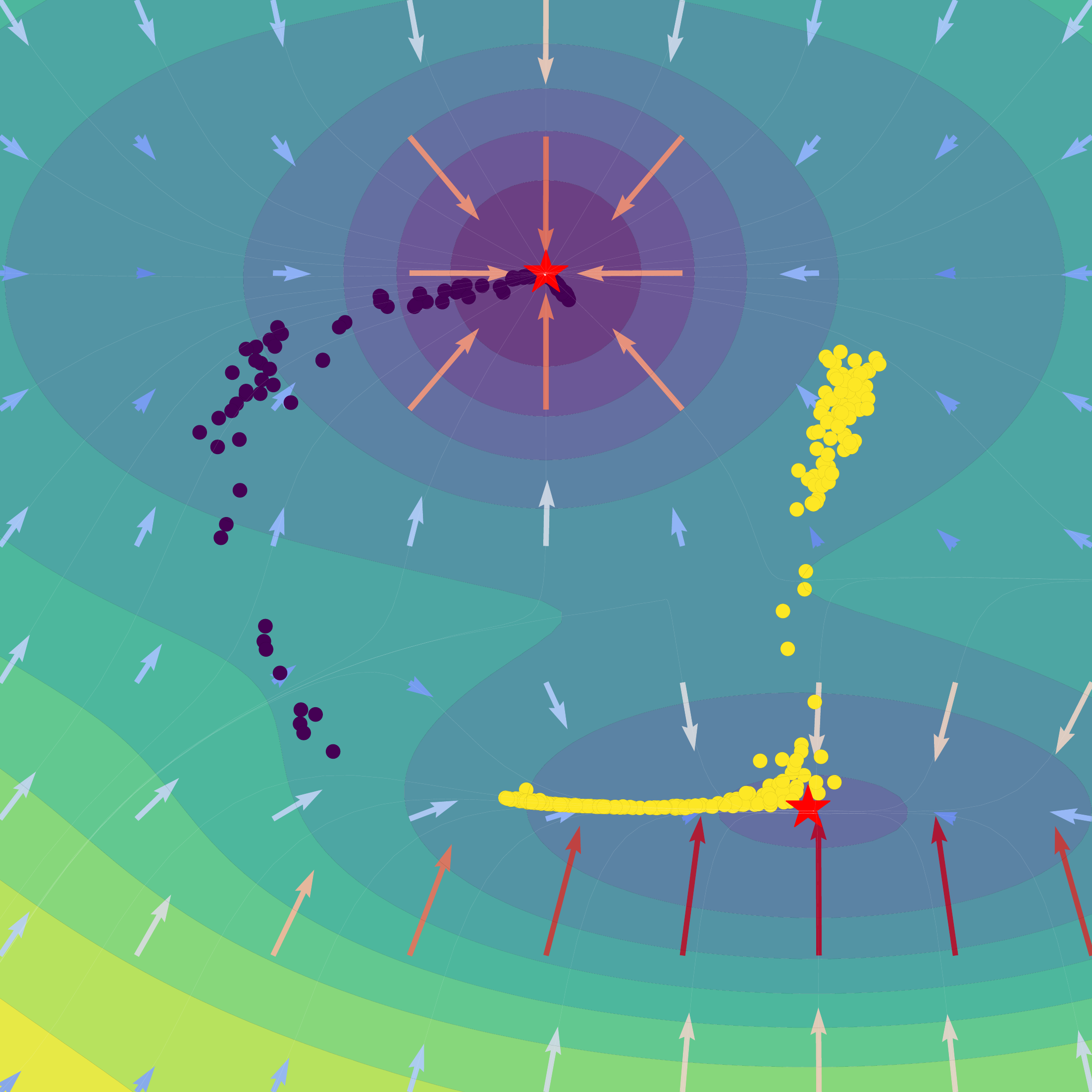}} 
    \end{tabular}
  }
  \end{minipage}
  \caption{Effect of regularization strength $\beta$ on a Gaussian SVFlow.
    (a): Synthetic 2D dataset, colored by class.
    (b): Training with only the variational consistency objective ($\beta\to\infty$).
    The vector field (arrows) is orthogonal to the ELBO contours (background), following the steepest descent direction.
    All points converge to a single ELBO optimum (red star), indicating posterior collapse onto a single Gaussian component.
    (c): Hybrid training with three regularization strengths.
    With no regularization ($\beta=0$), the flow separates classes well but the velocity field is not orthogonal to the ELBO contours and final states miss the ELBO optimum, indicating variational approximation error.
    As $\beta$ increases from $0.0$ to $0.5$, the field becomes more aligned with the ELBO gradient, improving probabilistic consistency but gradually reducing class separation.
  }
  \label{fig:beta_effect}
  \vspace{-4mm}
\end{figure*}

\textbf{Hybrid objective and the trade-off.}
The overall training loss combines the two terms:
\begin{equation}
\label{eq:hybrid_objective}
    \mathcal{J}_{\text{hybrid}}(\theta,\phi) = \mathcal{J}_\text{align}(\theta,\phi) + \beta\mathcal{J}_\text{var}(\theta,\phi),\quad\beta \geq 0.
\end{equation}
The semantic alignment term provides strong and well-conditioned gradients that prevent posterior collapse and encourage class‑conditional modes.
The variational consistency term acts as a regularizer, encouraging $q(z|x)\approx p(z|x)$ to align SVFlow vector field with the true score.
In practice, a small positive $\beta$ yields effective regularization without harming task performance,
whereas larger values can overwhelm the task signal and cause underfitting.
\autoref{fig:beta_effect} illustrates this balance on a 2D Gaussian SVFlow:
pure variational consistency collapses to a single mode, while hybrid training balances class separation and probabilistic consistency.

Notably, standard Transformer training with cross-entropy alone corresponds to $\beta=0$.
The architecture’s implicit biases, in particular the coupling between keys and queries in attention,
already provide a degree of variational consistency, which explains why Transformers learn well even without explicit regularization.
The gradient dynamic analysis in~\autoref{app:grad_derivation} reveals how the two objectives interact during optimization,
further elucidating their balancing mechanism.

\section{SVFlow Interpretation of Transformers}
\label{sec:transformer_flow}

\subsection{Spherical Manifold as Geometric Backbone}
We work on the unit sphere $\mathbb{S}^{d-1}$; the radius factor $\sqrt{d}$ in RMSNorm (Eq.~\eqref{eq:rmsnorm}) can be absorbed by scaling the state.
A Taylor expansion reveals that the residual connection and RMSNorm (Eq.~\eqref{eq:euler_step}) implement a \emph{relaxed retraction} to first order:
\begin{equation}
    \text{RMSNorm}(x+v) = R_x\left((I-xx^\top)v\right) + \mathcal{O}(\lVert v\rVert^2),\quad v\in\mathbb{R}^{d}
\end{equation}
This relaxed retraction allows us to apply Euclidean updates directly without explicit tangent projection, while still ensuring all latent states remain on the sphere,
providing a practical geometric backbone for flow discretization.

A natural distribution on the sphere is the von Mises-Fisher (vMF) distribution~\citep{wood1994simulation, arindam2005clustering}:
\begin{equation}
p_{\text{vMF}}(x;\mu, \kappa) = C_{d}(\kappa)\exp(\kappa\mu^\top x),
\end{equation}
where $\mu\in\mathbb{S}^{d-1}$ is the mean direction, $\kappa\in\mathbb{R}_{\geq 0}$ is the concentration,
and $C_d(\kappa) = \frac{\kappa^{d/2-1}}{(2\pi)^{d/2} I_{d/2-1}(\kappa)}$ is the normalizing constant with $I_\nu(\cdot)$ denoting the modified Bessel function of the first kind.
Its score in Euclidean space is linear in the mean direction, i.e., $\nabla_x\log p_{\text{vMF}}(x) = \kappa\mu$.
This linear score will be crucial for interpreting attention as a SVFlow.

\subsection{MHA as a vMF-based SVFlow}
We show that MHA $f_\text{Attn}$ (Eq.~\eqref{eq:mha}) implements a SVFlow vector field with vMF conditional distributions.

\textbf{Attention as Variational Posterior.}
For a single head $h$, treat each key $k_z$ as a latent component.
Define an unnormalized vMF distribution $\tilde{q}(x_q|z, h) = \exp\left(\kappa_z \mu_z^\top x_q\right)$ with $\kappa_z \mu_z = \left(W_{q,h}W_{k,h}^\top\right) k_z$.
Assuming a uniform prior $q(z) = 1/|\mathcal{Z}|$, the normalized variational posterior becomes 
\begin{equation}
  q(z|x_q, h)
    = \frac{\tilde{q}(x_q\vert z, h)q(z)}{\sum_{z^\prime\in\mathcal{Z}}\tilde{q}(x_q\vert z^\prime, h)q(z^\prime)}
    = \frac{\exp\left(\kappa_z \mu_z^\top x_q\right)}{\sum_{z'\in\mathcal{Z}} \exp\left(\kappa_{z'} \mu_{z'}^\top x_q\right)},
\end{equation}
which exactly matches the softmax attention weight $\sigma_{k_z,h}(x_q)$ in Eq.~\eqref{eq:attn_softmax}.
Thus, attention weights form a variational posterior over keys.

\begin{wrapfigure}{R}{0.5\textwidth}
  \centering
  \setlength{\tabcolsep}{3pt}
  \begin{tabular}{c c}
    \includegraphics[width=0.22\textwidth]{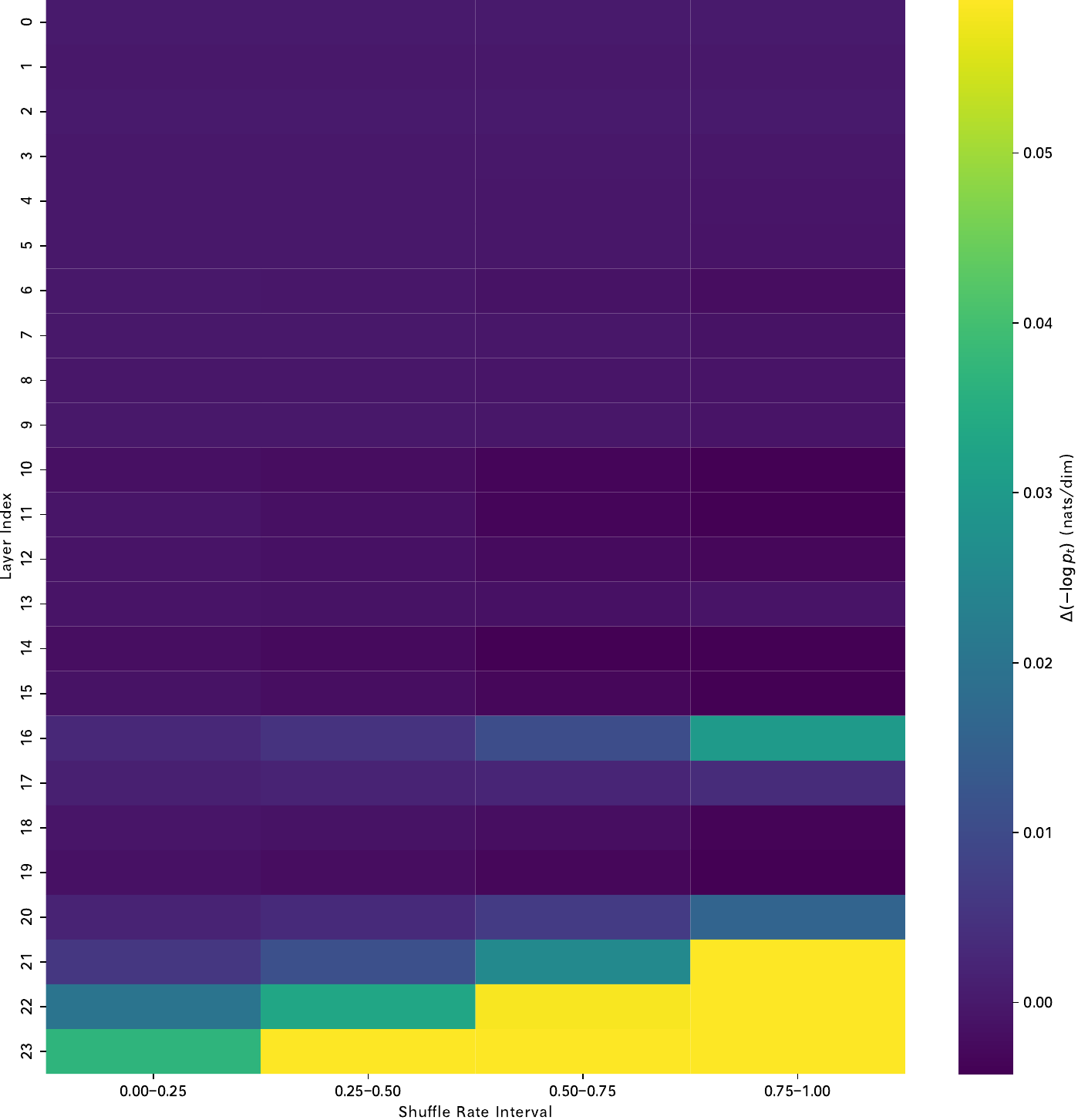} &
    \includegraphics[width=0.22\textwidth]{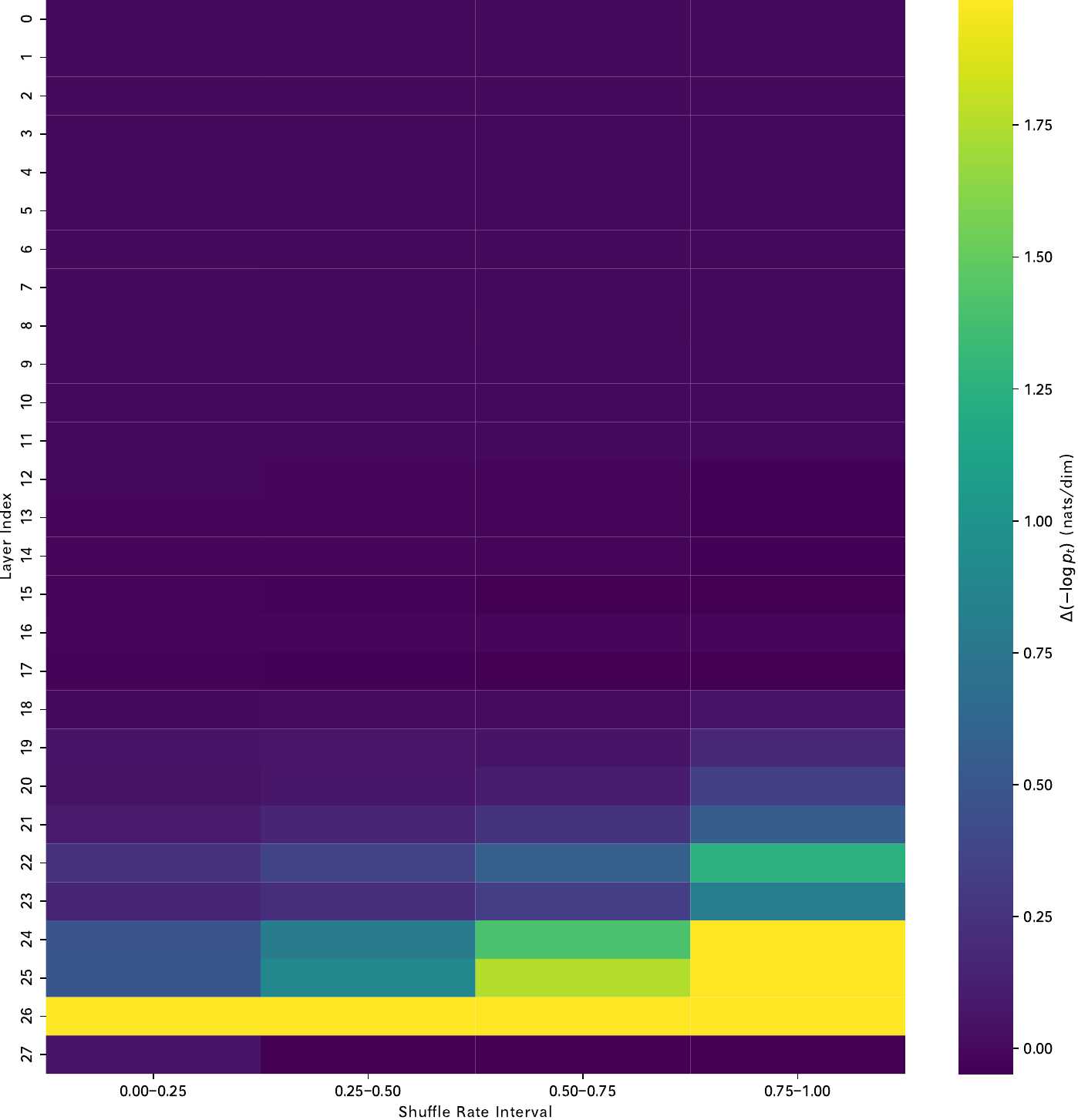} \\
    Qwen2.5-0.5B & Qwen3-0.6B
  \end{tabular}
  \caption{
    Layer-wise $\Delta(-\log p_t)$ for attention layers under different prefix shuffling rates.
    For both models, deep layers exhibit positive $\Delta$ while shallow layers remain near zero,
    indicating deeper attention representations are more sensitive to context disruption.
  }
  \label{fig:heatmap}
  \vspace{-6mm}
\end{wrapfigure}

\textbf{Value and Output Projection as Conditional Score.}
Similarly, let the conditional distribution $p(x|z)$ be vMF with parameters $\kappa_z^\prime\mu_z^\prime = \left(W_{o,h}W_{v,h}^\top\right) v_{z}$.
Its Euclidean score $\nabla_x \log p_\text{vMF}(x|z, h)=\kappa_z^\prime\mu_z^\prime$ is constant in $x$, matching the value‑output projection in Eq.~\eqref{eq:mha}.
Hence each value vector parameterizes a conditional score function.

\textbf{Head Output as SVFlow Vector Field.}
The head output is the posterior-weighted average of conditional scores:
\begin{equation}
v_{h}(x_q) = \sum_z q(z|x_q, h) \left(W_{o,h}W_{v,h}^\top\right) v_{z},
\end{equation}
which is precisely the SVFlow vector field for that head.
Treating head index as an additional latent variable, the full MHA output becomes an expectation over the joint distribution of heads and keys:
\begin{equation}
  f_{\text{Attn}}(x_q) = \mathbb{E}_{q(h\vert x_q)\,q(z\vert x_q,h)}\left[\nabla_x\log p_\text{vMF}(x_q|z,h)\right],
\end{equation}
where $q(z|x_q, h)$ and $\nabla_x\log p_\text{vMF}(x_q|z,h)$ are identified as above, $q(h\vert x_q)$ is input-dependent.
Standard MHA with uniform weighting corresponds to $q(h|x_q) = 1/H$;
recent gated head variants (e.g.,~\citet{csordas2024switchhead} and~\citet{qiu2025gatedattentionlargelanguage})
are naturally subsumed.

Notably, when the number of keys $\lvert\mathcal{Z}\rvert$ is large, the discrete posterior $q(z|x_q,h)$ converges to a continuous kernel‑smoothed posterior over the key distribution $p_{\text{key}}(k)$:
$q(k|x_q,h)\propto \exp\left((W_{q,h}W_{k,h}^\top k)^\top x_q\right) p_\text{key}(k)$, as detailed in Appendix~\ref{app:grad_implications}.
This aligns with the kernel smoother view of attention~\citep{tsai2019transformer} and, within our SVFlow framework,
reveals that MHA approximates the vector field $\mathbb{E}_{q(k|x_q,h)}[\nabla_x\log p_\text{vMF}(x|k,h)]$.
Consequently, each attention head implements a kernel‑based Monte Carlo estimate of the SVFlow dynamics, and the full MHA layer is a mixture of such fields.

\subsection{MoE as Network-based SVFlow Approximation}
The MoE layer $f_\text{MoE}$ (Eq.~\eqref{eq:moe}) admits an analogous interpretation.
Let experts be indexed by $z \in \{1, \dots, E\}$. In the SVFlow view:
(i) the gating network computes the variational posterior $q(z|x) = g_z(x)$;
(ii) each expert network approximates a conditional score $\nabla_x \log p(x\vert z) \approx e_z(x)$;
(iii) the MoE output is the SVFlow vector field $v_\text{MoE}(x) = f_{\text{MoE}}(x)$.
When $E=1$, this reduces to a standard FFN, which can be viewed as a single-component SVFlow.

\textbf{Relaxed Implementation.}
In a strict SVFlow, conditional scores obey the zero-mean property $\mathbb{E}_{p(x\vert z)}[\nabla_x\log p(x\vert z)] = 0$ under mild regularity conditions.
Standard MoE layers, however, do not enforce this constraint,
resulting in a relaxed SVFlow that trades exact probabilistic consistency for architectural flexibility.
Interestingly, recent work by~\citet{panda2026dense} uses exponential moving averages of expert outputs to center their distributions,
which can be viewed as an engineering approximation to the zero-mean constraint and has been shown to significantly improve training stability and performance.

\textbf{Expert Balancing as Variational Regularization.}
A common issue in MoE training is expert collapse, where only a few experts are active for most inputs.
The auxiliary load balancing loss~\citep{shazeer2017, fedus2022switch} is a widely used solution:
\begin{equation}
    \mathcal{J}_{\text{bal}} = \sum_{z=1}^E f_z \cdot P_z,\text{ with }
    f_z = \mathbb{E}_{p_\text{data}(x)}\left[ \mathds{1}\left\{\arg\max_{z^\prime} q(z^\prime\vert x) = z\right\}\right],
    P_z = \mathbb{E}_{p_\text{data}(x)}\left[q(z\vert x)\right].
\label{eq:moe_balancing}
\end{equation}
From the SVFlow perspective, this loss emerges as a coarse-grained variational regularizer, a view aligned with the theory of~\citet{su2026variational}.
Minimizing $\mathcal{J}_\text{bal}$ matches the marginal expert distribution to a uniform prior,
preventing posterior collapse at the population level while retaining per-sample flexibility.
Thus, $\mathcal{J}_\text{bal}$ can be viewed as a relaxation of the per-sample variational consistency objective $\mathcal{J}_\text{var}$:
While $\mathcal{J}_\text{var}$ enforces fine-grained probabilistic alignment, $\mathcal{J}_\text{bal}$ imposes a global marginal constraint.

\begin{figure*}[t]
  \vspace{-3mm}
  \centering
  \setlength{\tabcolsep}{3pt}
  \setlength\fboxsep{2pt}
  \begin{tabular}{c c c}
    \includegraphics[width=0.32\textwidth]{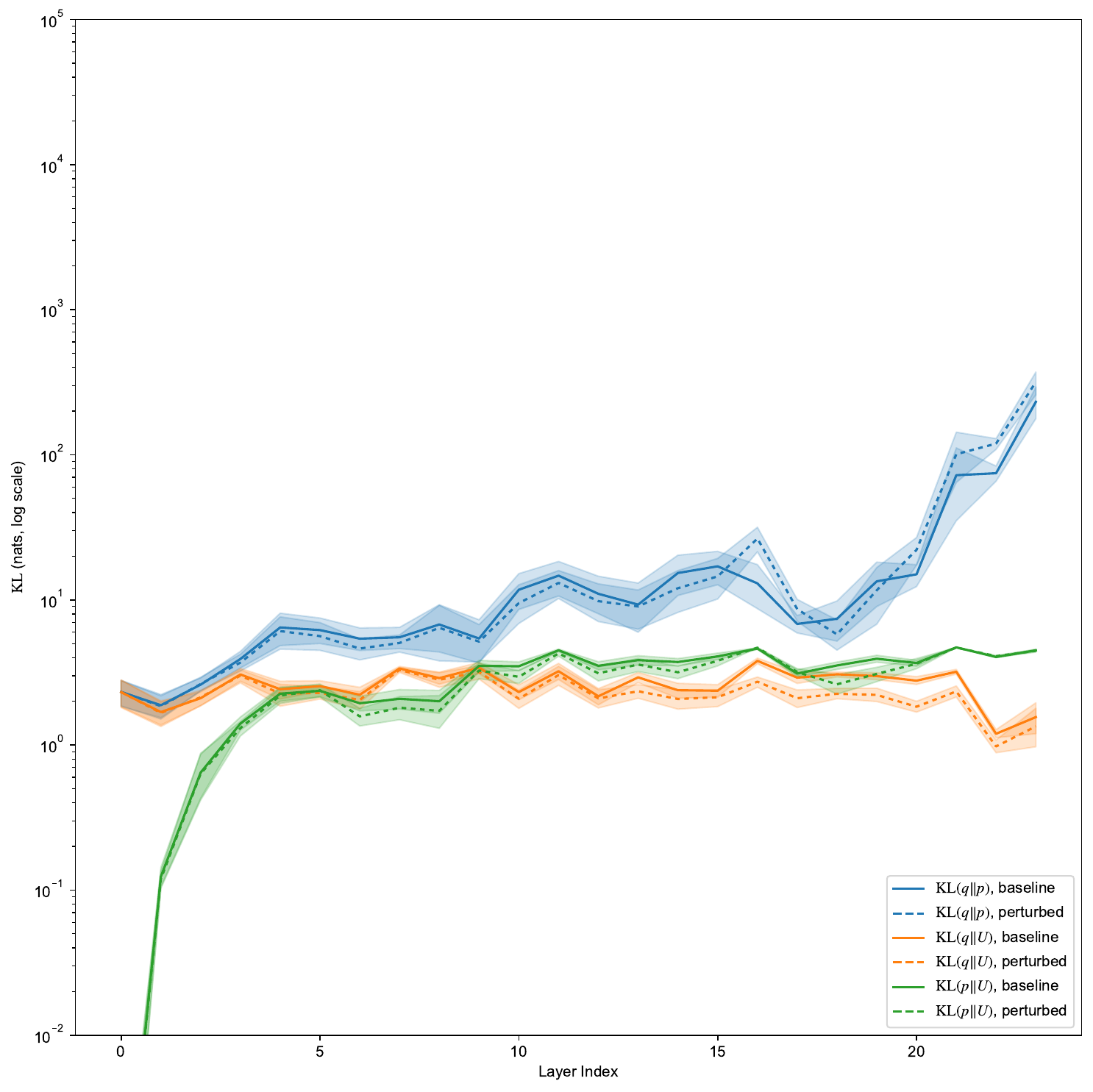} &
    \includegraphics[width=0.32\textwidth]{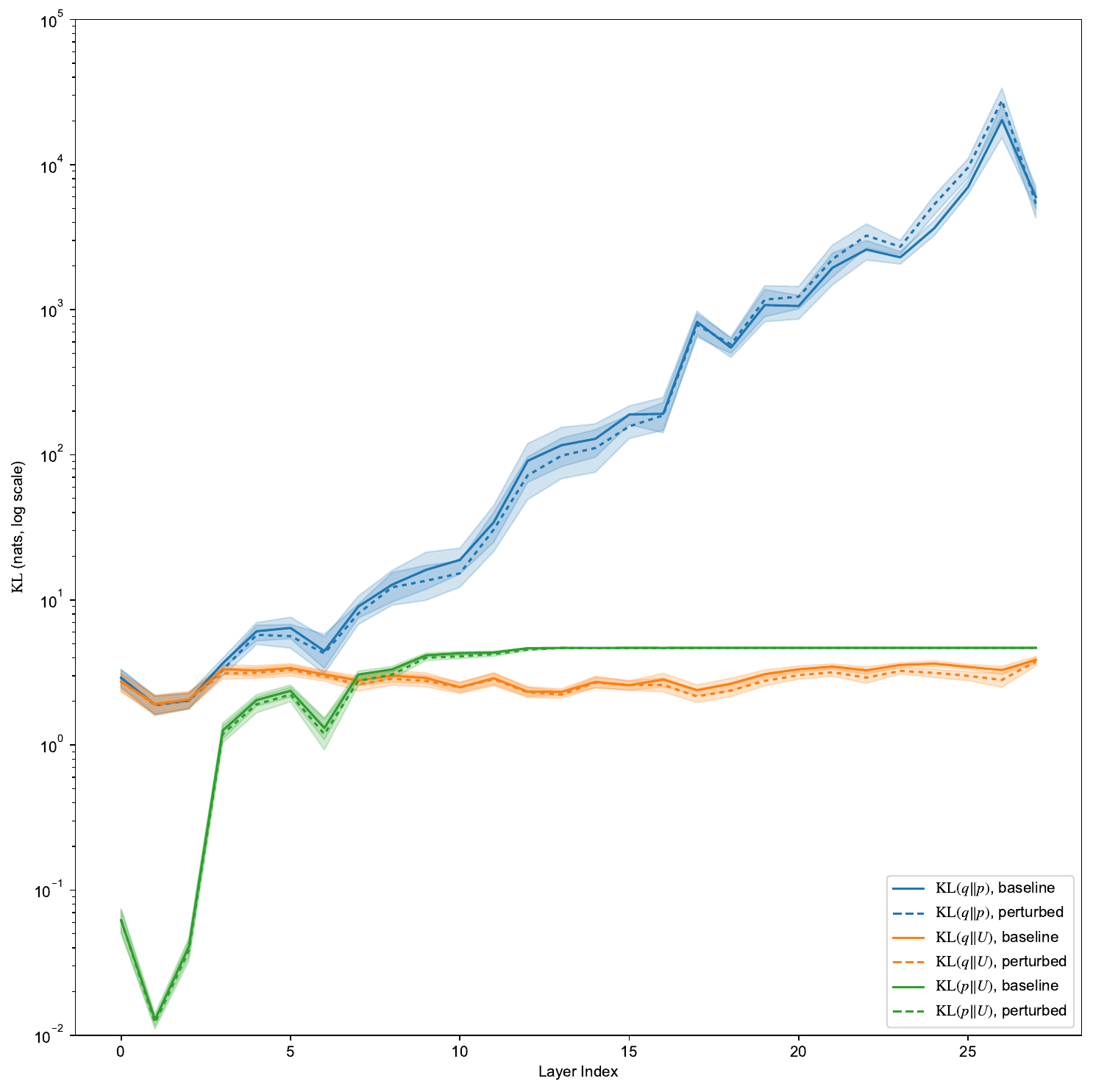} &
    \includegraphics[width=0.32\textwidth]{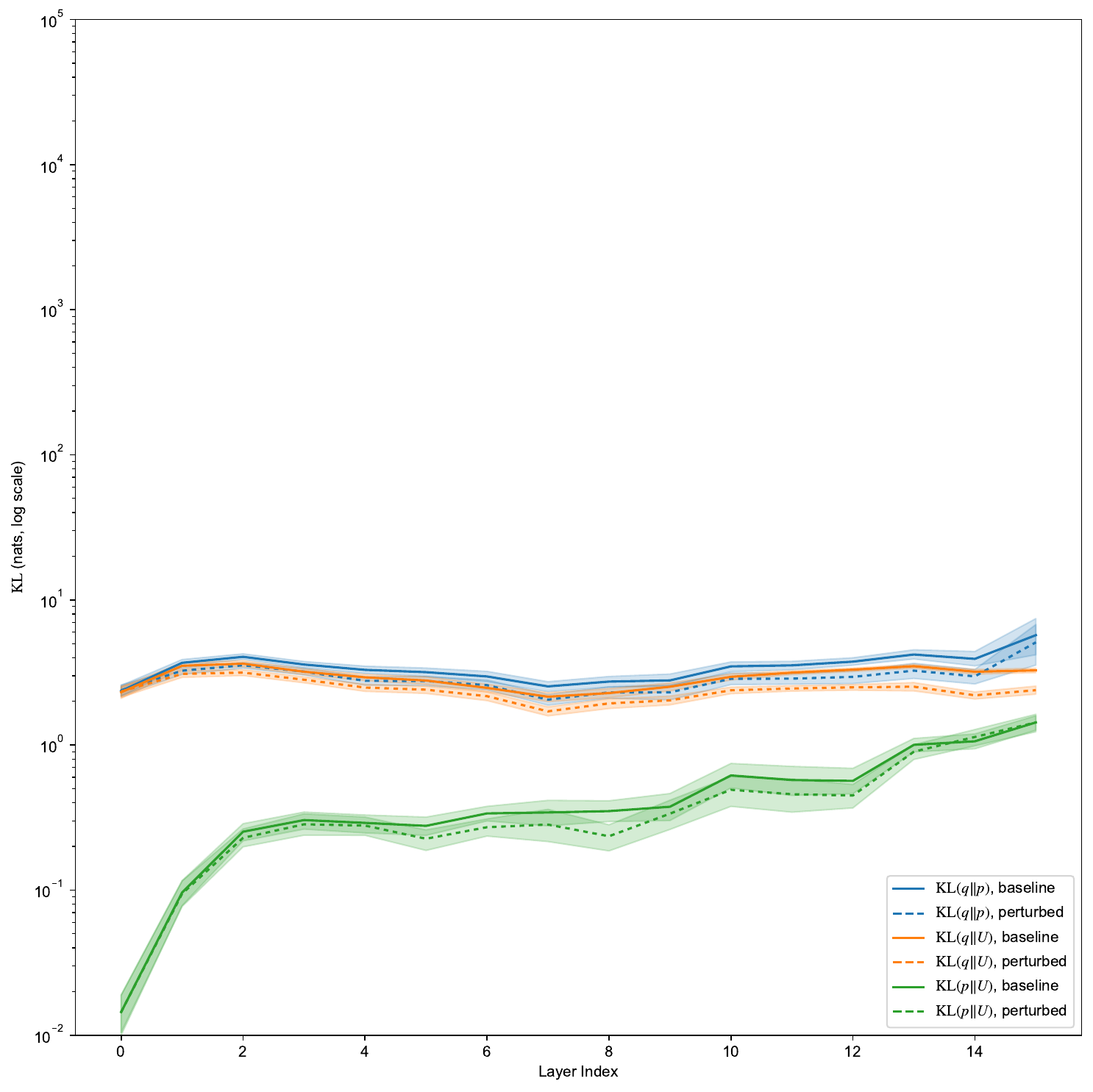} \\
    Qwen2.5-0.5B & Qwen3-0.6B & Llama3.2-1B
  \end{tabular}
  \caption{
      Layer‑wise evolution of divergence $\mathrm{KL}(q\Vert p)$ and concentrations $\mathrm{KL}(q\Vert U)$ and $\mathrm{KL}(p\Vert U)$
      under baseline (solid) and highest shuffle rate (dashed) conditions. The $y$-axis is logarithmic.
      While $\mathrm{KL}(q\Vert U)$ remains in a similar moderate range across all three models, $\mathrm{KL}(p\Vert U)$ differs dramatically:
      it stays low and flat for Llama3.2, rises gradually for Qwen2.5, and surges at early layers for Qwen3.
      Consequently, $\mathrm{KL}(q\Vert p)$ remains stable in Llama3.2, rises mildly in Qwen2.5, and explodes in Qwen3,
      corresponding respectively to the low-sensitivity, moderate, and high‑sensitivity regimes.
  }
  \label{fig:kl_plot}
  \vspace{-3mm}
\end{figure*}

\textbf{Why MHA Does Not Require Explicit Regularization.}
\emph{Unlike MoE, the key–value coupling in MHA balances the positive feedback in variational posterior gradients.}
As derived in Appendix~\ref{app:grad_implications}, components favored by the task loss receive stronger gradients, creating a positive feedback loop that drives expert collapse in MoE.
In MHA, however, the same representation that generates the key parameterizing variational posterior also produces the value defining conditional score;
gradients from the score branch flow back into the shared representation, providing a balancing force that mitigates the feedback.
This implicit regularization makes MHA generally more robust and often trains stably without auxiliary load‑balancing losses,
though extreme collapse (e.g., attention sink~\citep{xiao2024efficient}) may still occur.
Such regularization has been verified in simplified settings by recent studies on self-attention~\citep{tarzanagh2023transformers, vasudeva2025implicit, sheen2024implicit}.

\subsection{Transformer as Discrete Spherical SVFlow}
A Transformer block integrates the preceding components into a forward Euler discretization of the spherical SVFlow ODE.
Each layer applies either attention or MoE as the vector field $v_\ell$,
where the discrete layer index $\ell$ acts as time and the layer‑wise independent parameters simulate a fixed‑step evolution.
This is followed by spherical retraction via residual connection and RMSNorm.
The token embeddings serve as the initial state $x_0$, drawn from the empirical token distribution $p_0$,
and the final layer's output $x_T = x^L$ feeds into a classification head to produce $p_T(y\vert x_T)$.
Under the standard cross-entropy objective ($\beta=0$), the flow drives representations toward class-conditional modes,
yielding linearly separable latent states for discriminative learning.

Thus, SVFlow provides a unified probabilistic and geometric lens through which Transformer architectures can be understood and analyzed.
The theoretical insights derived from this interpretation are empirically investigated in Section~\ref{sec:experiments}.

\section{Related Work}
\label{sec:related}
\textbf{Generative vs. Discriminative Flows.}
Neural ODEs~\citep{chen2018neural} established the view of residual networks as discretizations of continuous dynamics.
Score-based generative models~\citep{song2021scorebased} showed that an ODE driven by the score function can transport distributions.
Flow matching~\citep{lipman2023flow} and variational flow matching~\citep{eijkelboom2024variational} provide simulation-free training.
Recent convergence analyses~\citep{huang2025convergence} further solidify the theory,
and large-scale Wasserstein gradient flows~\citep{mokrov2021large} connect to optimal transport.
Unlike these \emph{generative} flows, SVFlow is a \emph{discriminative} representation flow that drives representations toward class-conditional modes,
a distinction critical for understanding Transformer representations.

\textbf{Probabilistic and Variational Interpretations.}
Several works link attention to probabilistic inference.
\citet{tsai2019transformer} view attention as kernel smoothing.
\citet{hu2022fuse} propose a variational Transformer with layer-wise latent variables for text generation.
\citet{ravuri2025transformers} interpret Transformers as unrolled inference in a probabilistic Laplacian eigenmap.
\citet{xu2018spherical} introduce von Mises‑Fisher distributions for spherical latent spaces.
\citet{desai2020calibration} empirically study calibration of pre‑trained Transformers.
SVFlow extends these partial views by formalizing the entire forward pass as a discrete ODE where the vector field is a variational posterior-weighted average of conditional scores.

\textbf{Dynamics, Geometry, and Implicit Regularization.}
A line of work interprets Transformer layers through dynamical systems~\citep{geshkovski2023emergence,geshkovski2025math} and geometry.
Continuous ODE interpretations have been proposed~\citep{yang2026unifying,tai2025mathematical,li2021ode,tong2025neural}.
\citet{gupta2025layernorm} analyzes RMSNorm from a geometric perspective.
Implicit regularization in attention has been studied via gradient dynamics~\citep{agarwal2026gradient,sheen2024implicit,vasudeva2025implicit,tarzanagh2023transformers}.
SVFlow integrates these insights by treating the residual-RMSNorm combination as a relaxed retraction,
and key‑value coupling as implicit variational regularization.

\textbf{Mixture of Experts Theory.}
MoE layers~\citep{shazeer2017,fedus2022switch} are regularized by heuristic load‑balancing losses.
\citet{guo2025advancing} shows that this auxiliary loss can inadvertently lead to expert overlap and hinder specialization, highlighting its limitations.
\citet{su2026variational} provides a unifying variational interpretation, showing that the balancing loss matches the marginal expert distribution to a uniform prior.
\citet{panda2026dense} shows that dense gradients to the router improve MoE training.
\citet{cai2025asurvey} provide a comprehensive survey.
SVFlow unifies these insights by interpreting the gating network as a variational posterior and each expert as a conditional score approximator.
The load-balancing loss emerges as a coarse-grained version of our variational consistency objective.

\begin{table*}[t]
  \vspace{-3mm}
  \centering
  \small
  \caption{Deep‑layer aggregated metrics for three pre‑trained models under prefix shuffling.
    For each shuffle rate interval, we report the baseline value and the change $\Delta$ (perturbed minus baseline).
    Units: $-\log p_t$ in nats/dim; KL divergences and logPPL in nats; ECE in \%.
  }
  \label{tab_prefix_shuffle}
  \setlength{\tabcolsep}{1.6pt}
  \begin{tabular}{lccccccccccccc}
    \toprule
      \multirow{2}{*}{Model} & \multirow{2}{*}{Shuffle Rate} & \multicolumn{2}{c}{$-\log p_t$} & \multicolumn{2}{c}{$\mathrm{KL}(q\|p)$} &
      \multicolumn{2}{c}{$\mathrm{KL}(q\|U)$} & \multicolumn{2}{c|}{$\mathrm{KL}(p\|U)$} & \multicolumn{2}{c}{ECE (\%)} & \multicolumn{2}{c}{logPPL} \\
    \cmidrule(lr){3-4}\cmidrule(lr){5-6}\cmidrule(lr){7-8}\cmidrule(lr){9-10}\cmidrule(lr){11-12}\cmidrule(lr){13-14}
    & & Base & $\Delta$ & Base & $\Delta$ & Base & $\Delta$ & Base & $\Delta$ & Base & $\Delta$ & Base & $\Delta$ \\
    \midrule
    \multirow{4}{*}{Qwen2.5}
      & 0.00-0.25 & -1.892 & $3.74e^{-3}$  & 34.551 &  1.814 & 2.887 & -0.082 & 4.007 & -0.142 & 0.70 & 1.19 & 2.70 & 0.07 \\
      & 0.25-0.50 & -1.895 & $6.33e^{-3}$  & 33.210 &  3.005 & 2.740 & -0.123 & 3.847 & -0.174 & 0.75 & 2.56 & 2.80 & 0.19 \\
      & 0.50-0.75 & -1.896 & $11.74e^{-3}$ & 32.855 &  5.449 & 2.700 & -0.209 & 3.802 & -0.220 & 0.81 & 5.29 & 2.81 & 0.50 \\
      & 0.75-1.00 & -1.896 & $22.87e^{-3}$ & 32.854 & 10.391 & 2.699 & -0.475 & 3.799 & -0.276 & 0.70 & 6.91 & 2.80 & 1.78 \\
    \cmidrule{1-14}
    \multirow{4}{*}{Qwen3}
      & 0.00-0.25 & 3.488 & $0.202$ & 2616.72 & 108.11 & 3.211 & -0.046 & 4.914 & -0.025 & 9.24 & -1.41 & 3.23 & 0.03 \\
      & 0.25-0.50 & 3.353 & $0.321$ & 2547.43 & 169.16 & 3.047 & -0.063 & 4.668 & -0.028 & 9.46 & -2.14 & 3.34 & 0.14 \\
      & 0.50-0.75 & 3.318 & $0.540$ & 2529.19 & 280.37 & 3.004 & -0.088 & 4.606 & -0.032 & 9.50 & -3.16 & 3.37 & 0.40 \\
      & 0.75-1.00 & 3.312 & $1.281$ & 2526.77 & 659.44 & 3.003 & -0.220 & 4.606 & -0.038 & 9.38 & -6.69 & 3.35 & 1.37 \\
    \cmidrule{1-14}
    \multirow{4}{*}{Llama3.2}
      & 0.00-0.25 & -2.369 & $-3.35e^{-4}$ & 3.748 & -0.332 & 3.075 & -0.333 & 0.646 & -0.008 & 0.49 & 1.42 & 2.43 & 0.09 \\
      & 0.25-0.50 & -2.368 & $-3.55e^{-4}$ & 3.573 & -0.348 & 2.912 & -0.360 & 0.634 & -0.016 & 0.41 & 2.27 & 2.54 & 0.18 \\
      & 0.50-0.75 & -2.368 & $-4.01e^{-4}$ & 3.527 & -0.390 & 2.869 & -0.410 & 0.631 & -0.034 & 0.39 & 4.18 & 2.56 & 0.42 \\
      & 0.75-1.00 & -2.368 & $-6.39e^{-4}$ & 3.529 & -0.620 & 2.871 & -0.627 & 0.630 & -0.065 & 0.44 & 6.74 & 2.54 & 1.72 \\
    \bottomrule
  \end{tabular}
  \vspace{-3mm}
\end{table*}

\section{Experiments}
\label{sec:experiments}
To evaluate the SVFlow interpretation of Transformer, we conduct prefix‑shuffling experiments for attention layers on pre‑trained language models.

\subsection{Experimental Setup}
We test on three models: Qwen2.5‑0.5B, Qwen3‑0.6B and Llama3.2‑1B.
Using the Wikitext‑2 test set,
we generate perturbed versions of each sequence by randomly permuting the first $\lfloor pN\rfloor$ tokens with $p\in\{0.2,0.4,0.6,0.8,0.95\}$;
the unperturbed sequence with $p=0$ serves as baseline.
For a token at position $n$, its prefix shuffle rate is $r=\min(1, pN/n)$.
Tokens with $n > pN$ are used and grouped into four bins by $r$.

For each token we compute the marginal negative log-likelihood $-\log p_t(x)$, the variational divergence $\mathrm{KL}(q\|p)$,
and the concentrations $\mathrm{KL}(q\|U)$ and $\mathrm{KL}(p\|U)$ for uniform $U$,
together with log perplexity (logPPL) and expected calibration error (ECE).
See Appendix~\ref{app_pretrained_model} for details.

Because deep layers dominate the sensitivity to perturbation (\autoref{fig:heatmap}),
we aggregate over the deepest third of layers for the main analysis.
\autoref{tab_prefix_shuffle} reports these deep‑layer aggregated metrics, including baseline values and the change under perturbation.
Full‑layer averages are in Appendix~\autoref{tab_full}.

\subsection{Marginal Likelihood and Task Performance}
We examine how hidden representation quality, measured by $-\log p_t$, relates to task performance logPPL and ECE.

\textbf{Baseline correlation.} From the baseline columns of~\autoref{tab_prefix_shuffle}, $-\log p_t$ and logPPL show a clear monotonic trend across models:
Llama3.2 has the lowest $-\log p_t$ and the lowest logPPL; Qwen3 has the highest $-\log p_t$ and the highest logPPL; Qwen2.5 lies in between.
Over the 12 data points (3 models $\times$ 4 bins), the Spearman correlation between $-\log p_t$ and logPPL is $\rho=0.87$ with $p=2\times 10^{-4}$, 
indicating that higher marginal likelihood implies better predictive performance.
A similarly strong correlation holds for ECE ($\rho=0.86$, $p=4\times10^{-4}$), confirming that higher marginal likelihood also implies better calibration.

\textbf{Changes under perturbation.} Under prefix shuffling, Qwen2.5 and Qwen3 exhibit positive $\Delta(-\log p_t)$ and $\Delta$logPPL that increase with the shuffle rate,
consistent with context disruption degrading both representation and prediction.
For Llama3.2, however, $\Delta(-\log p_t)$ remains near zero while $\Delta$logPPL rises sharply.
As we show next, the perturbation effects on calibration follow similarly distinct patterns, which we analyze through concentration and divergence dynamics.

\subsection{Concentration-Divergence Analysis of Calibration}
We investigate how calibration depends on the concentrations $\mathrm{KL}(q\|U)$, $\mathrm{KL}(p\|U)$ and the divergence $\mathrm{KL}(q\|p)$,
under baseline and perturbation with $r\in(0.75, 1.0)$.

As derived in Appendix~\ref{app:concentration_and_divergence}, low concentrations bound $\mathrm{KL}(q|p)$ while high concentrations make it sensitive to mismatches.
Appendix~\autoref{fig:kappa} shows the layer‑wise vMF concentration $\kappa$ (determining $\mathrm{KL}(p\|U)$) for each model.
The orders of magnitude differ dramatically: Llama3.2 reaches $\sim10^2$, Qwen2.5 reaches $\sim10^3$, and Qwen3 reaches $\sim10^5$.
Perturbation leaves these orders unchanged, so the sharpness of $p$ is unaffected by shuffling.
This $\kappa$ disparity determines the three sensitivity regimes below.

\textbf{Moderate regime (Qwen2.5).} Baseline concentrations are moderate and close, $\mathrm{KL}(q\|p)$ is modest, and ECE is low.
Perturbation slightly decreases concentrations, raises $KL(q\|p)$, and worsens ECE.
As shown in~\autoref{fig:kl_plot} (left), concentrations stay moderate while $\mathrm{KL}(q\|p)$ gradually increases in deep layers,
reflecting mild sensitivity.

\textbf{High‑sensitivity regime (Qwen3).} Baseline $\kappa$ is extreme, making $\mathrm{KL}(p\|U)$ much larger than $\mathrm{KL}(q\|U)$; $\mathrm{KL}(q\|p)$ is huge and ECE is high.
Perturbation slightly reduces $\mathrm{KL}(q\|U)$, further increases $\mathrm{KL}(q\|p)$, yet improves calibration.
This occurs because baseline $p$ is over‑confident: even small misalignment inflates $\mathrm{KL}(q\|p)$,
but the misaligned $q$ spreads mass more evenly, avoiding the incorrect dominant component that caused high ECE.
\autoref{fig:kl_plot} (middle) shows $\mathrm{KL}(p\|U)$ rising sharply then saturating,
while $\mathrm{KL}(q\|U)$ fluctuates without increase.
The resulting gap makes $\mathrm{KL}(q\|p)$ explode in deep layers, a hallmark of high sensitivity.

\textbf{Low-sensitivity regime (Llama3.2).} Baseline $\kappa$ stays low, so $\mathrm{KL}(p\|U)$ is extremely low while $\mathrm{KL}(q\|U)$ is moderate, $\mathrm{KL}(q\|p)$ is also low and ECE is excellent.
Perturbation leaves $\mathrm{KL}(q\|p)$ nearly unchanged but ECE rises sharply.
\autoref{fig:kl_plot} (right) shows all quantities low and flat across layers,
indicating that attention‑level KL divergences no longer dominate calibration.

In conclusion, concentrations determine the sensitivity of divergence.
When moderate and balanced, divergence $\mathrm{KL}(q\|p)$ reliably tracks calibration,
so penalizing it via $\mathcal{J}_\text{var}$ may help prevent over‑confidence.

\section{Conclusion}
We introduced SVFlow and showed that Transformer is an approximation of discretized spherical SVFlow.
Under the framework, we characterized the relation between SVFlow metrics and task performance, providing insight on the underlying mechanism of attention.
Overall, SVFlow offers a principled foundation for analyzing and designing Transformers.
A limitation is the reliance on pre-trained models; future work should train SVFlow from scratch.

\bibliographystyle{plain}
\bibliography{refs}

\medskip

\newpage
\appendix

\section{Proofs}

\subsection{SVFlow Evidence Lower Bound}
\label{proof:elbo_bound}
We provide the detailed derivation of the instantaneous ELBO for SVFlow, following the standard variational inference argument applied pointwise along the continuous trajectory.
For any fixed time $t$, the marginal log-density $\log p_t(x)$ admits the decomposition:
\begin{align}
  \log p_t(x) &= \int q_t(z\vert x)\log p_t(x) dz \\
            &= \int q_t(z\vert x) \log \frac{p_t(x, z)}{p_t(z\vert x)} dz \\
            &= \int q_t(z\vert x) \left[\log\frac{p_t(x, z)}{q_t(z\vert x)} + \log\frac{q_t(z\vert x)}{p_t(z\vert x)}\right] dz \\
            &= \underbrace{\int q_t(z\vert x) \log\frac{p_t(x, z)}{q_t(z\vert x)}dz}_{\mathcal{L}_t(x)} +
               \underbrace{\int q_t(z\vert x)\log\frac{q_t(z\vert x)}{p_t(z\vert x)}dz}_{\mathrm{KL}(q_t(z\vert x)\Vert p_t(z\vert x)) \ge 0}.
\end{align}
The inequality $\log p_t(x) \ge \mathcal{L}_t(x)$ holds pointwise, with equality iff $q_t(z|x)=p_t(z|x)$.
This pointwise variational interpretation directly motivates the training objectives discussed in the main text, in particular the variational consistency regularization $\mathcal{J}_\text{var}$.

\subsection{Gradient decomposition}
\label{proof:grad_decomp}
We derive the gradient decomposition $\nabla_x \mathcal{L}_t(x) = v_t(x) + \epsilon_t(x)$ from Theorem~\ref{thm:grad_decomp}:
\begin{align}
  \nabla_x{\mathcal{L}_{t}(x)} 
    &= \nabla_x\int q_t(z\vert x)\log\frac{p_t(x, z)}{q_t(z\vert x)}dz \\
    &= \int\nabla_x \left[q_t(z\vert x)\log\frac{p_t(x, z)}{q_t(z\vert x)}\right] dz \\
    &= \int\log\frac{p_t(x, z)}{q_t(z\vert x)} \nabla_x q_t(z\vert x) dz +
       \int q_t(z\vert x) \nabla_x \log p_t(x, z) dz -
       \cancel{\int q_t(z\vert x) \nabla_x \log {q_t(z\vert x)} dz} \label{eq_grad_decomp_tag1}\\
    &= \int\left[\log\frac{p_t(z\vert x)}{q_t(z\vert x)}+\log p_t(x)\right] \nabla_x q_t(z\vert x) dz +
       \int q_t(z\vert x)\left[\nabla_x\log p_t(x\vert z) + \cancel{\nabla_x\log p_t(z)}\right] dz \\
    &= \int\log\frac{p_t(z\vert x)}{q_t(z\vert x)} \nabla_x q_t(z\vert x) dz +
       \cancel{\int\log p_t(x)\nabla_x q_t(z\vert x) dz} +
       \int q_t(z\vert x) \nabla_x \log p_t(x\vert z) dz \label{eq_grad_decomp_tag2} \\
    &= \underbrace{\int\log\frac{p_t(z\vert x)}{q_t(z\vert x)} \nabla_x q_t(z\vert x) dz}_{\epsilon_t(x)} +
       \underbrace{\int q_t(z\vert x) \nabla_x \log p_t(x\vert z) dz}_{v_t(x)},
\end{align}
where step~\eqref{eq_grad_decomp_tag1} uses the identity $\nabla_x q_t(z|x) = q_t(z|x)\nabla_x\log q_t(z|x)$, thus the third term vanishes because:
\begin{equation}
\int q_t(z\vert x) \nabla_x \log {q_t(z\vert x)} dz = \int \nabla_x q_t(z\vert x) dz = \nabla_x \int q_t(z\vert x) dz = \nabla_x 1 = 0.
\end{equation}
Similarly, the second term in step~\eqref{eq_grad_decomp_tag2} vanishes because:
\begin{equation}
    \int \log p_t(x) \nabla_x \log {q_t(z\vert x)} dz = \log p_t(x) \int \nabla_x q_t(z\vert x) dz = \log p_t(x)\cdot \nabla_x 1 = 0.
\end{equation}

This completes the derivation of Eq.~\eqref{eq:grad_decomp}.

\section{Gradient Dynamics of SVFlow Training}
\label{app:grad_derivation}
We provide a complete gradient derivation for the discrete-time SVFlow training objective, 
followed by a discussion that connects the results to the main text claims on the posterior collapse of pure variational consistency,
the implicit regularization in MHA, and the kernel smoothing interpretation of attention.

\subsection{Setup}
Consider the flow where the latent state evolves as 
\begin{equation}
\label{eq_residual}
  x^{\ell+1} = x^\ell + \sum_{z\in\mathcal{Z}}q_{\phi_\ell}(z\vert x^\ell)s_{\theta_\ell}(x^\ell, z), \quad \ell = 0, 1, \cdots, L-1,
\end{equation}
with $x^0$ being the initial state and $x^L=x_T$ the final state used for classification via
$p_T(y\vert x^L; W) = \text{softmax}(W^\top x^L)_y$.
The variational posterior is defined as $q_{\phi_\ell}(z\vert x)=\text{softmax}(g_{\phi_\ell}(x))_z$ with logits $g_{\phi_\ell}(x)$,
and the conditional score function is $s_{\theta_\ell}(x, z)=\nabla_x\log p_{\theta_\ell}(x\vert z)$.
For a single sample $(x_0, y)$ and assuming uniform time-weighting $\lambda(t)$ for simplicity,
the hybrid training objective is
\begin{equation}
    \mathcal{J}_\text{hybrid} = \underbrace{-\log p(y\vert x^L)}_{\mathcal{J}_\text{align}} +
    \beta\sum_{\ell=0}^{L-1}\underbrace{\mathrm{KL}(q_{\phi_\ell}(z\vert x^\ell)\Vert p_{\theta_\ell}(z\vert x^\ell))}_{\mathcal{J}_\text{var}^\ell}.
\end{equation}
Define the classification error signal $\delta_\ell = \partial\mathcal{J}_\text{align}/\partial{x^\ell}$ backpropagated from the final layer,
and the variational error signal $\gamma_{k,\ell}=\partial\mathcal{J}_\text{var}^k/\partial{x^\ell}$ for downstream layers $k>\ell$.

\subsection{Gradient for Posterior Parameters $\phi_\ell$}
By the chain rule, the gradient decomposes into three contributions:
\begin{align}
  \frac{\partial\mathcal{J}_\text{hybrid}}{\partial\phi_\ell} &= \frac{\partial\mathcal{J}_\text{align}}{\partial\phi_\ell} +
  \beta\frac{\partial\mathcal{J}_\text{var}^\ell}{\partial\phi_\ell} +
  \beta\sum_{k=\ell+1}^{L-1} \frac{\partial\mathcal{J}_\text{var}^k}{\partial\phi_\ell} \label{eq_grad_phi_1} \\
  &= \delta_{\ell+1}^\top\frac{\partial x^{\ell+1}}{\partial\phi_\ell} +
  \beta\frac{\partial\mathcal{J}_\text{var}^\ell}{\partial \phi_\ell} +
    \beta\sum_{k=\ell+1}^{L-1}\gamma_{k,\ell+1}^\top\frac{\partial x^{\ell+1}}{\partial\phi_\ell}. \label{eq_grad_phi_2}
\end{align}
The gradient from $\mathcal{J}_\text{var}^{<\ell}$ vanishes since $\phi_\ell$ does not affect earlier layers.

For the first term, from the update equation~\eqref{eq_residual} and noting that $s_{\theta_\ell}$ does not depend on $\phi_\ell$, we have:
\begin{align}
  \delta_{\ell+1}^\top\frac{\partial x^{\ell+1}}{\partial\phi_\ell}
    &= \delta_{\ell+1}^\top\sum_z s_{\theta_\ell}(x^\ell, z)\cdot \left(\frac{\partial q_{\phi_\ell}(z\vert x^\ell)}{\partial\phi_\ell}\right)^\top \\
    &= \sum_z \left(\delta_{\ell+1}^\top s_{\theta_\ell}(x^\ell, z)\right)\cdot\frac{\partial q_{\phi_\ell}(z\vert x^\ell)}{\partial\phi_\ell} \\
    &= \sum_z A_\ell(z) \frac{\partial q_{\phi_\ell}(z\vert x^\ell)}{\partial\phi_\ell}, \label{eq_grad_phi_delta}
\end{align}
where $A_\ell(z) = \delta_{\ell+1}^\top s_{\theta_\ell}(x^\ell, z)\in\mathbb{R}$ is the \emph{task advantage function}.

For the third term, following the same derivation, we have:
\begin{equation}
\label{eq_grad_phi_gamma}
  \gamma_{k,\ell+1}^\top\frac{\partial x^{\ell+1}}{\partial\phi_\ell}
    = \sum_z B_{k,\ell}(z) \frac{\partial q_{\phi_\ell}(z\vert x^\ell)}{\partial\phi_\ell},
\end{equation}
where $B_{k,\ell} = \gamma_{k,\ell+1}^\top s_{\theta_\ell}(x^\ell, z)\in\mathbb{R}$ is the \emph{future variational advantage function}.

For the second term, following from the derivative of KL divergence, we have: 
\begin{align}
  \frac{\partial\mathcal{J}_\text{var}^\ell}{\partial \phi_\ell}
    &= \frac{\partial}{\partial\phi_\ell}\sum_z q_{\phi_\ell}(z\vert x^\ell)\log\frac{q_{\phi_\ell}(z\vert x^\ell)}{p_{\theta_\ell}(z\vert x^\ell)} \\
    &= \sum_z\frac{\partial q_{\phi_\ell}(z\vert x^\ell)}{\partial\phi_\ell}\log\frac{q_{\phi_\ell}(z\vert x^\ell)}{p_{\theta_\ell}(z\vert x^\ell)}
    + \cancel{\sum_z q_{\phi_\ell}(z\vert x^\ell)\frac{\partial}{\partial\phi_\ell}\log q_{\phi_\ell}(z\vert x^\ell)}
    - \cancel{\sum_z q_{\phi_\ell}(z\vert x^\ell)\frac{\partial}{\partial\phi_\ell}\log p_{\theta_\ell}(z\vert x^\ell)} \\
    &= \sum_z R_\ell(z) \frac{\partial q_{\phi_\ell}(z\vert x^\ell)}{\partial\phi_\ell},  \label{eq_grad_phi_local_gamma}
\end{align}
where $R_\ell(z) = \log\frac{q_{\phi_\ell}(z\vert x^\ell)}{p_{\theta_\ell}(z\vert x^\ell)}$ is the \emph{self variational advantage function}.

Notice that~\eqref{eq_grad_phi_delta}, ~\eqref{eq_grad_phi_gamma} and~\eqref{eq_grad_phi_local_gamma} all are the summation of posterior gradient w.r.t $\phi_\ell$ but with different advantage weightings.
Using the gradient of softmax operation:
\begin{equation}
  \frac{\partial q_{\phi_\ell}(z\vert x^\ell)}{\partial\phi_\ell} =
  q_{\phi_\ell}(z\vert x^\ell)\sum_{z^\prime}(\mathds{1}_{z^\prime=z} - q_{\phi_\ell}(z^\prime\vert x^\ell))\frac{\partial g_{\phi_\ell}(x^\ell)_{z^\prime}}{\partial\phi_\ell},
\end{equation}
and taking the term~\eqref{eq_grad_phi_delta} as an example, we obtain:
\begin{align}
  \delta_{\ell+1}^\top\frac{\partial x^{\ell+1}}{\partial\phi_\ell}
    &= \sum_z A_\ell(z)\cdot q_{\phi_\ell}(z\vert x^\ell)\sum_{z^\prime}(\mathds{1}_{z^\prime=z} - q_{\phi_\ell}(z^\prime\vert x^\ell))\frac{\partial g_{\phi_\ell}(x^\ell)_{z^\prime}}{\partial\phi_\ell} \\
    &= \sum_{z^\prime}\left[\sum_{z}\left[\left(\mathds{1}_{z^\prime=z} - q_{\phi_\ell}(z^\prime\vert x^\ell)\right) q_{\phi_\ell}(z\vert x^\ell)A_\ell(z)\right]\right]\frac{\partial g_{\phi_\ell}(x^\ell)_{z^\prime}}{\partial\phi_\ell} \label{eq_grad_q_phi_1} \\
    &= \sum_{z^\prime}\left[\sum_{z}\left[\mathds{1}_{z^\prime=z} q_{\phi_\ell}(z\vert x^\ell)A_\ell(z)\right] - \sum_z\left[q_{\phi_\ell}(z^\prime\vert x^\ell)q_{\phi_\ell}(z\vert x^\ell)A_\ell(z)\right]\right]\frac{\partial g_{\phi_\ell}(x^\ell)_{z^\prime}}{\partial\phi_\ell} \label{eq_grad_q_phi_2} \\
    &= \sum_{z^\prime}\left[q_{\phi_\ell}(z^\prime\vert x^\ell)A_\ell(z^\prime) - q_{\phi_\ell}(z^\prime\vert x^\ell)\mathbb{E}_{q_{\phi_\ell}(z\vert x^\ell)}\left[A_\ell(z)\right]\right]\frac{\partial g_{\phi_\ell}(x^\ell)_{z^\prime}}{\partial\phi_\ell} \\
    &= \sum_{z^\prime}q_{\phi_\ell}(z^\prime\vert x^\ell)\left[A_\ell(z^\prime) - \mathbb{E}_{q_{\phi_\ell}(z\vert x^\ell)}\left[A_\ell(z)\right]\right]\frac{\partial g_{\phi_\ell}(x^\ell)_{z^\prime}}{\partial\phi_\ell}.
\end{align}
Step~\eqref{eq_grad_q_phi_1} exchanges the summation between $z$ and $z^\prime$,
and step~\eqref{eq_grad_q_phi_2} expands the product inside the inner summation to separate the indicator term from the probability product term,
which allows the simplification using the definition of expectation.
Applying the same algebraic manipulation to~\eqref{eq_grad_phi_gamma} and~\eqref{eq_grad_phi_local_gamma} yields analogous expressions,
with $A_\ell$ replaced by $B_{k,\ell}$ and $R_\ell$ respectively.

Assembling all contributions gives the final gradient for the posterior parameters:
\begin{equation}
    \frac{\partial\mathcal{J}_{\text{hybrid}}}{\partial\phi_\ell} = \sum_z q_{\phi_\ell}(z|x^\ell)\eta_{\ell}(z) \frac{\partial g_{\phi_\ell}(x^\ell)_z}{\partial\phi_\ell},
\end{equation}
where the \emph{total advantage function} $\eta_\ell(z)\in\mathbb{R}$ aggregates the three centered forces:
\begin{equation}
\label{eq_grad_advantage_eta}
    \eta_{\ell}(z) = \Big(A_\ell(z) - \mathbb{E}_{q_{\phi_\ell}(z^\prime\vert x^\ell)}\left[A_\ell(z^\prime)\right]\Big) +
    \beta \Big(R_\ell(z) - \mathbb{E}_{q_{\phi_\ell}(z^\prime\vert x^\ell)}\left[R_\ell(z^\prime)\right]\Big) +
    \beta \sum_{k=\ell+1}^{L-1}\Big(B_{k,\ell}(z) - \mathbb{E}_{q_{\phi_\ell}(z^\prime\vert x^\ell)}\left[B_{k,\ell}(z^\prime)\right]\Big).
\end{equation}
The centering (subtracting the mean) of advantage functions ensures that the update redistributes probability mass among components rather than shifting all logits uniformly.

\subsection{Gradient for Score Parameters $\theta_\ell$}
The gradient for $\theta_\ell$ involves two types of dependence: through the score function $s_{\theta_\ell}$ in the forward pass and
through the conditional densities $p_{\theta_\ell}(z\vert x^\ell)$ in the local variational consistency $\mathcal{J}_\text{var}^\ell$.

By the chain rule, the gradient decomposes similarly:
\begin{align}
\frac{\partial\mathcal{J}_{\text{hybrid}}}{\partial\theta_\ell}
  &= \frac{\partial\mathcal{J}_{\text{align}}}{\partial\theta_\ell}
    + \beta\frac{\partial\mathcal{J}_{\text{var}}^\ell}{\partial\theta_\ell}
    + \beta\sum_{k=\ell+1}^{L-1} \frac{\partial\mathcal{J}_{\text{var}}^k}{\partial\theta_\ell} \\
  &= \delta_{\ell+1}^\top\frac{\partial x^{\ell+1}}{\partial\theta_\ell} +
  \beta\frac{\partial\mathcal{J}_\text{var}^\ell}{\partial \theta_\ell} +
    \beta\sum_{k=\ell+1}^{L-1}\gamma_{k,\ell+1}^\top\frac{\partial x^{\ell+1}}{\partial\theta_\ell}. \label{eq_grad_theta_2}
\end{align}

From the update equation~\eqref{eq_residual}:
\begin{align}
\frac{\partial x^{\ell+1}}{\partial\theta_\ell} = \sum_z q_{\phi_\ell}(z|x^\ell) \frac{\partial s_{\theta_\ell}(x^\ell, z)}{\partial\theta_\ell},
\end{align}
and the derivative of local variational consistency:
\begin{align}
  \frac{\partial\mathcal{J}_\text{var}^\ell}{\partial \theta_\ell}
    &= \frac{\partial}{\partial\theta_\ell}\sum_z q_{\phi_\ell}(z\vert x^\ell)\log\frac{q_{\phi_\ell}(z\vert x^\ell)}{p_{\theta_\ell}(z\vert x^\ell)} \\
    &= -\sum_z q_{\phi_\ell}(z\vert x^\ell) \frac{\partial}{\partial\theta_\ell} \log p_{\theta_\ell}(z\vert x^\ell) \\
    &= -\sum_z q_{\phi_\ell}(z\vert x^\ell) \frac{\partial}{\partial\theta_\ell} \log \frac{p_{\theta_\ell}(x^\ell\vert z) p(z)}{\sum_{z^\prime} p_{\theta_\ell}(x^\ell\vert z^\prime) p(z^\prime)} \\
    &= -\sum_z q_{\phi_\ell}(z\vert x^\ell) \left(\frac{\partial}{\partial\theta_\ell} \log p_{\theta_\ell}(x^\ell\vert z) - \frac{\partial}{\partial\theta_\ell}\log{\sum_{z^\prime} p_{\theta_\ell}(x^\ell\vert z^\prime) p(z^\prime)}\right) \\
    &= -\sum_z q_{\phi_\ell}(z\vert x^\ell) \left(\frac{\partial}{\partial\theta_\ell} \log p_{\theta_\ell}(x^\ell\vert z) - \sum_{z^\prime}{p_{\theta_\ell}(z^\prime\vert x^\ell)}\left[\frac{\partial}{\partial\theta_\ell}\log p_{\theta_\ell}(x^\ell\vert z^\prime)\right]\right) \\
    &= -\sum_z q_{\phi_\ell}(z\vert x^\ell)\frac{\partial}{\partial\theta_\ell} \log p_{\theta_\ell}(x^\ell\vert z) + \sum_z q_{\phi_\ell}(z\vert x^\ell)\left[\sum_{z^\prime}{p_{\theta_\ell}(z^\prime\vert x^\ell)}\left[\frac{\partial}{\partial\theta_\ell}\log p_{\theta_\ell}(x^\ell\vert z^\prime)\right]\right] \\
    &= -\sum_z q_{\phi_\ell}(z\vert x^\ell)\frac{\partial}{\partial\theta_\ell} \log p_{\theta_\ell}(x^\ell\vert z) + \sum_{z^\prime}\left(\sum_z q_{\phi_\ell}(z\vert x^\ell)\right){p_{\theta_\ell}(z^\prime\vert x^\ell)}\left[\frac{\partial}{\partial\theta_\ell}\log p_{\theta_\ell}(x^\ell\vert z^\prime)\right] \\
    &= \sum_z \left[p_{\theta_\ell}(z\vert x^\ell) - q_{\phi_\ell}(z\vert x^\ell)\right]\frac{\partial}{\partial\theta_\ell} \log p_{\theta_\ell}(x^\ell\vert z).
\end{align}
Thus, the final gradient for the score parameters is:
\begin{equation}
\label{eq_grad_score_param}
  \frac{\partial\mathcal{J}_{\text{hybrid}}}{\partial\theta_\ell} =
    \sum_z q_{\phi_\ell}(z\vert x^\ell) \left( \delta_{\ell+1}^\top + \beta\sum_{k=\ell+1}^{L-1} \gamma_{k,\ell+1}^\top \right) \frac{\partial s_{\theta_\ell}(x^\ell, z)}{\partial\theta_\ell} +
    \beta \sum_z \big[ p_{\theta_\ell}(z\vert x^\ell) - q_{\phi_\ell}(z\vert x^\ell) \big] \frac{\partial}{\partial\theta_\ell} \log p_{\theta_\ell}(x^\ell|z).
\end{equation}
The first term aligns the score functions with both task and downstream variational objectives,
while the second term (which vanishes when $q_{\phi_\ell} = p_{\theta_\ell}$) adjusts the conditional densities to improve local variational consistency.

\subsection{Implications for Transformer Training Dynamics}
\label{app:grad_implications}
The gradient expressions derived above provide analytical support for three key claims in the main text.
We discuss each in turn, emphasizing the role of task supervision, kernel smoothing, and parameter sharing.

\paragraph{Collapse under Pure Variational Consistency ($\beta\to\infty$).}
When the task loss is absent (or $\beta\to\infty$, thus task term is negligible), the advantage function~\eqref{eq_grad_advantage_eta} reduces to
\begin{equation}
  \eta_\ell(z) = \beta \left(R_\ell(z)-\mathbb{E}_q[R_\ell]\right) + \beta \sum_{k=\ell+1}^{L-1}\left(B_{k,\ell}(z)-\mathbb{E}_q[B_{k,\ell}]\right).
\end{equation}
The centering operation of self variational advantage $R_\ell(z)$ drives $q$ toward $p$, but creates a positive feedback loop:
components with above-average advantage $R_\ell(z) > \mathbb{E}_q[R_\ell]$ receive positive updates and become even more favored, while others are suppressed.
For the last layer ($\ell = L-1$), the feedback relies solely on centered $R_{L-1}(z)$, leading to a winner-takes-all situation.
For earlier layers ($\ell \le L-1$), future advantages $B_{k,\ell}(z)$ propagate positive feedback from downstream layers, amplifying the imbalance from top to bottom.
Thus, without a task signal to counteract this loop, the posterior quickly collapses to a single component.
This explains why pure \(\mathcal{J}_\text{var}\) minimization in mixture models fails and why a task supervision loss (e.g., cross‑entropy) is essential in practice.

\paragraph{Kernel Smoothing Limit of Attention.}
Consider a single attention head with keys $\{k_z\}_{z\in\mathcal{Z}}$ drawn i.i.d. from an empirical distribution $p_{\text{key}}$.
For a fixed query $x_q$, the attention weight is
\begin{equation}
q(z\vert x_q) = \frac{\exp(\kappa_z\mu_z^\top x_q)}{\sum_{j=1}^N \exp(\kappa_j\mu_j^\top x_q)},
\end{equation}
where $\kappa_z\mu_z = W_qW_k^\top k_z$, and $N=\lvert Z\rvert$. Define the kernel function $\psi(x_q, k) = \exp((W_qW_k^\top k)^\top x_q)$.
By the law of large numbers, as $N\to\infty$,
\begin{equation}
\frac{1}{N}\sum_{j=1}^N \psi(x_q, k_j) \to \mathbb{E}_{k\sim p_{\text{key}}}[\psi(x_q, k)] \quad \text{almost surely}.
\end{equation}
For any test function $f$, the weighted sum converges weakly to a continuous integral:
\begin{equation}
    \sum_{z=1}^N f(k_z)q(z\vert x_q) \to \frac{\int f(k) \psi(x_q, k) p_{\text{key}}(k) dk}{\int \psi(x_q, k') p_{\text{key}}(k') dk'}.
\end{equation}
Thus, the discrete posterior $q(z\vert x_q)$ converges to the continuous kernel‑smoothed posterior
\begin{equation}
q(k|x_q) = \frac{\psi(x_q, k) p_{\text{key}}(k)}{\int \psi(x_q, k') p_{\text{key}}(k') dk'}.
\end{equation}
This result follows directly from the strong law of large numbers and the continuous mapping theorem; it requires only that $\psi(x_q, \cdot)$ is bounded (or integrable) with respect to $p_{\text{key}}$, which holds for the exponential kernel.

This kernel view not only justifies the vMF mixture interpretation but also sets the stage for understanding how the task loss implicitly regularizes the key distribution, as discussed next.

\paragraph{Why MHA Does Not Require Explicit Regularization.}
When only the task loss $\mathcal{J}_\text{align}$ is used ($\beta=0$), the posterior parameters $\phi_\ell$ (keys) receive gradient only through the centered task advantage $A_\ell(z)-\mathbb{E}_q[A_\ell]$.
This term alone would create a positive feedback loop similar to the pure variational consistency case.
However, in MHA the keys and values are derived from the same input representation via different projections.
This shared representation receives gradients from both the posterior branch (through $\nabla_{\phi_\ell} \mathcal{J}_\text{align}$)
and the score branch (through $\nabla_{\theta_\ell}\mathcal{J}_{\text{align}}$).
The score branch gradient flows into the shared representation and provides a balancing force:
when a key becomes too dominant (large $q(k_z\vert x_q)$), the corresponding value’s contribution to the task loss tends to saturate or produce opposing gradients,
because the value is also updated via the same representation.
Crucially, even if a key has low attention weight (i.e., $q(k_z\vert x_q)\approx0$), its corresponding token representation can still receive strong gradients from other roles:
as a query attending to other keys, or as a value for other queries.
From the kernel viewpoint, the task loss implicitly shapes the key distribution $p_{\text{key}}$ to be non‑degenerate on the data manifold, ensuring the posterior to remain non‑collapsing.
In MoE, by contrast, each expert’s parameters appear only in its own output; the gating network and experts are decoupled.
If an expert’s gating probability $g_i$ is small, its output is suppressed, and its parameters receive negligible gradients, leading to “expert death”.
Hence, MoE requires explicit auxiliary load‑balancing losses, whereas MHA’s multi‑role coupling provides implicit regularization.

\section{Experimental Details}

\subsection{Synthetic 2D Experiment}
To illustrate the trade‑off between semantic alignment and variational consistency, we train a Gaussian SVFlow on 2D datasets using the hybrid objective with different $\beta$.

\paragraph{Dataset and Model.}
We generate synthetic moons (noise 0.06) dataset using scikit-learn.
For each run, points are randomly sampled.
The SVFlow uses $8$ mixture components, i.e., $\mathcal{Z}=\{1, 2, \dots, 8\}$.
The variational posterior $q(z\vert x) = \text{softmax}(log q(x\vert z))$ and
conditional likelihood $p(x\vert z)$ are diagonal Gaussians with time‑dependent parameters predicted by a time‑embedding network (3-layer MLP, hidden size 64).
The ODE is integrated with forward Euler (step size $0.01$).
Training uses batch size 512, Adam (learning rate 0.01, beta1 0.9, beta2 0.999, no weight decay), and 10,000 steps for all configurations.
The hybrid objective (Eq.~\eqref{eq:hybrid_objective}) is evaluated with $\beta\in \{0.0, 0.1, 0.5\}$; the pure variational consistency case ($\beta\to\infty$)
is approximated by training only with $\mathcal{J}_\text{var}$. ELBO and $\mathcal{J}_\text{var}$ are computed via closed‑form Gaussians.

\subsection{Prefix Shuffling Experiments}
\label{app_pretrained_model}
\subsubsection{models and data}
We evaluate pre-trained Large Language Models (LLMs) in inference mode: Qwen2.5-0.5B-Instruct (24 layers), Qwen3‑0.6B (28 layers), and Llama3.2‑1B (16 layers).
The Wikitext‑2 test set is used and preprocessed by removing samples shorter than 50 characters, yielding 1,945 sequences.

\subsubsection{prefix shuffling and binning}
For each sequence of length $N$, we generate perturbed versions by randomly permuting the first $\lfloor pN\rfloor$ tokens with shuffle proportion
$p\in\{0, 0.2, 0.4, 0.6, 0.8, 0.95\}$; the suffix remains unchanged. The unperturbed sequence ($p=0$) serves as the baseline.
For a token at position $n$ (1-indexed), its prefix shuffle rate is $r=\min(1, pN / n)$.
Tokens with $n \le pN$ (fully shuffled) are excluded;
Tokens with $n > pN$ are grouped into four bins by $r$: $(0.0, 0.25]$, $(0.25, 0.5]$, $(0.5, 0.75]$, $(0.75, 1.0)$.

\subsubsection{metrics and relative deviation}
For each token pair (perturbed vs. baseline), we compute per layer and per head metrics using a modified attention implementation in Hugging Face Transformers.
From these we obtain:
\begin{itemize}
  \item marginal negative log‑likelihood $-\log p_t = -\mathcal{L}_t - \mathrm{KL}(q_t\Vert p_t)$ (nats/dim),
  \item variational divergence $\mathrm{KL}(q_t\Vert p_t)$ (nats),
  \item variational concentration $\mathrm{KL}(q\vert U)$ (nats),
  \item true concentration $\mathrm{KL}(p\|U)$ (nats).
\end{itemize}
The change $\Delta$ for each metric $m$ is the value on perturbed tokens minus that on baseline tokens, i.e., $\Delta m = m_\text{pert} - m_\text{base}$.
For task performance we compute log-Perplexity (logPPL, nats) and expected calibration error (ECE, 15 bins) for each bin.

Because deep layers dominate the sensitivity (Fig.~\ref{fig:heatmap}),
we aggregate over the \emph{deepest third of layers} (last 8 layers for Qwen2.5, last 9 for Qwen3, last 5 for Llama3.2) for the main table.
Full‑layer averages are given in Table~\ref{tab_full} in next.

\textbf{Numerical stability.} For the vMF distribution, the log‑normalizer $\log C_d(\kappa)$ involves the modified Bessel function $I_{d/2-1}(\kappa)$.
For large $\kappa$ (e.g., $\kappa>10^3$ in deep layers of Qwen3), evaluating $\log C_d(\kappa)$ can overflow.
Consequently, the computed $\log p(x|z)$ and thus $\mathrm{KL}(p\|U)$ may become unreliable in those layers.
We retain the reported values as indicative of a trend (extremely high concentration) but caution that absolute numbers may be affected by numerical precision.
The key qualitative conclusion, a large and persistent gap between $\mathrm{KL}(p\|U)$ and $\mathrm{KL}(q\|U)$,
remains robust because even a lower bound on $\mathrm{KL}(p\|U)$ would still exceed $\mathrm{KL}(q\|U)$.

\subsubsection{relationship between concentration and divergence}
\label{app:concentration_and_divergence}
Let $\mathcal{Z}$ be a finite set with $|\mathcal{Z}|$ elements and $U$ the uniform distribution over $\mathcal{Z}$.
For any distributions $q$ and $p$ on $\mathcal{Z}$, we have:
\begin{align}
  \mathrm{KL}(q\|U) &= \sum_z q(z)\log\frac{q(z)}{1/|\mathcal{Z}|} = \log|\mathcal{Z}| - H(q), \\
  \mathrm{KL}(p\|U) &= \log|\mathcal{Z}| - H(p), \\
  \mathrm{KL}(q\|p) &= \sum_z q(z)\log\frac{q(z)}{p(z)} = H(q,p) - H(q),
\end{align}
where $H(q)=-\sum_z q(z)\log q(z)$ is the entropy and $H(q,p)=-\sum_z q(z)\log p(z)$ the cross entropy.

When both $q$ and $p$ are close to uniform, $H(q)\approx H(p)\approx \log|\mathcal{Z}|$ and $H(q,p)\approx \log|\mathcal{Z}|$ (because $p$ is nearly uniform).
Hence $\mathrm{KL}(q\|p)$ is bounded by a small constant.

When both $q$ and $p$ are highly concentrated, their entropies are small.
However, the cross entropy $H(q,p)$ can be large if $q$ and $p$ put their mass on different elements.
Since $H(q,p)$ has no upper bound (it can approach $+\infty$ as $p(z)\to 0$ for the elements where $q(z)>0$), $\mathrm{KL}(q\|p)$ can become arbitrarily large.

Therefore, the concentration measures $\mathrm{KL}(q\|U)$ and $\mathrm{KL}(p\|U)$ control the sensitivity of $\mathrm{KL}(q\|p)$ to mismatches:
low concentration guarantees a bounded divergence, while high concentration makes divergence potentially explosive.

\subsubsection{additional results}
\label{app:shuffling_results}
Table~\ref{tab_full} reports the same metrics averaged over \emph{all layers}.
Compared to the deep‑layer aggregates, the full‑layer values show smaller $\Delta$ because shallow layers contribute little change, confirming that the observed trends are driven by deep layers.

\begin{table*}[htpb]
  \centering
  \caption{
    Full‑layer averages (all layers) for the three models.
  }
  \label{tab_full}
  \setlength{\tabcolsep}{1.6pt}
  \begin{tabular}{lccccccccc}
    \toprule
      \multirow{2}{*}{Model} & \multirow{2}{*}{Shuffle Rate} & \multicolumn{2}{c}{$-\log p_t$ (nats/dim)} & \multicolumn{2}{c}{$\mathrm{KL}(q\|p)$} & \multicolumn{2}{c}{$\mathrm{KL}(q\|U)$} & \multicolumn{2}{c}{$\mathrm{KL}(p\|U)$} \\
    \cmidrule(lr){3-4}\cmidrule(lr){5-6}\cmidrule(lr){7-8}\cmidrule{9-10}
    & & Base & $\Delta$ & Base & $\Delta$ & Base & $\Delta$ & Base & $\Delta$ \\
    \midrule
    \multirow{4}{*}{Qwen2.5-0.5B}
      & 0.00-0.25 & -1.917 & $2.48e^{-3}$  & 24.540 & 1.204 & 2.812 & -0.062 & 3.135 & -0.103 \\
      & 0.25-0.50 & -1.919 & $4.17e^{-3}$  & 23.584 & 1.987 & 2.662 & -0.094 & 3.019 & -0.129 \\
      & 0.50-0.75 & -1.919 & $7.72e^{-3}$  & 23.332 & 3.590 & 2.622 & -0.158 & 2.986 & -0.167 \\
      & 0.75-1.00 & -1.919 & $15.04e^{-3}$ & 23.332 & 6.827 & 2.621 & -0.347 & 2.983 & -0.218 \\
    \cmidrule{1-10}
    \multirow{4}{*}{Qwen3-0.6B}
      & 0.00-0.25 & 1.718 & 0.137 & 1777.46 &  73.37 & 3.158 & -0.036 & 3.834 & -0.025 \\
      & 0.25-0.50 & 1.626 & 0.218 & 1730.38 & 114.78 & 2.994 & -0.052 & 3.653 & -0.031 \\
      & 0.50-0.75 & 1.602 & 0.367 & 1717.99 & 190.22 & 2.951 & -0.075 & 3.607 & -0.042 \\
      & 0.75-1.00 & 1.599 & 0.869 & 1716.35 & 447.36 & 2.950 & -0.180 & 3.606 & -0.062 \\
    \cmidrule{1-10}
    \multirow{4}{*}{Llama3.2-1B}
      & 0.00-0.25 & -2.369 & $-3.19e^{-4}$ & 3.708 & -0.320 & 3.160 & -0.321 & 0.504 & -0.005 \\
      & 0.25-0.50 & -2.368 & $-3.38e^{-4}$ & 3.533 & -0.337 & 2.991 & -0.344 & 0.496 & -0.011 \\
      & 0.50-0.75 & -2.368 & $-3.76e^{-4}$ & 3.487 & -0.373 & 2.947 & -0.383 & 0.494 & -0.025 \\
      & 0.75-1.00 & -2.368 & $-5.47e^{-4}$ & 3.488 & -0.540 & 2.948 & -0.537 & 0.493 & -0.048 \\
    \bottomrule
  \end{tabular}
\end{table*}

\begin{figure}[ht]
  \centering
  \includegraphics[width=0.5\textwidth]{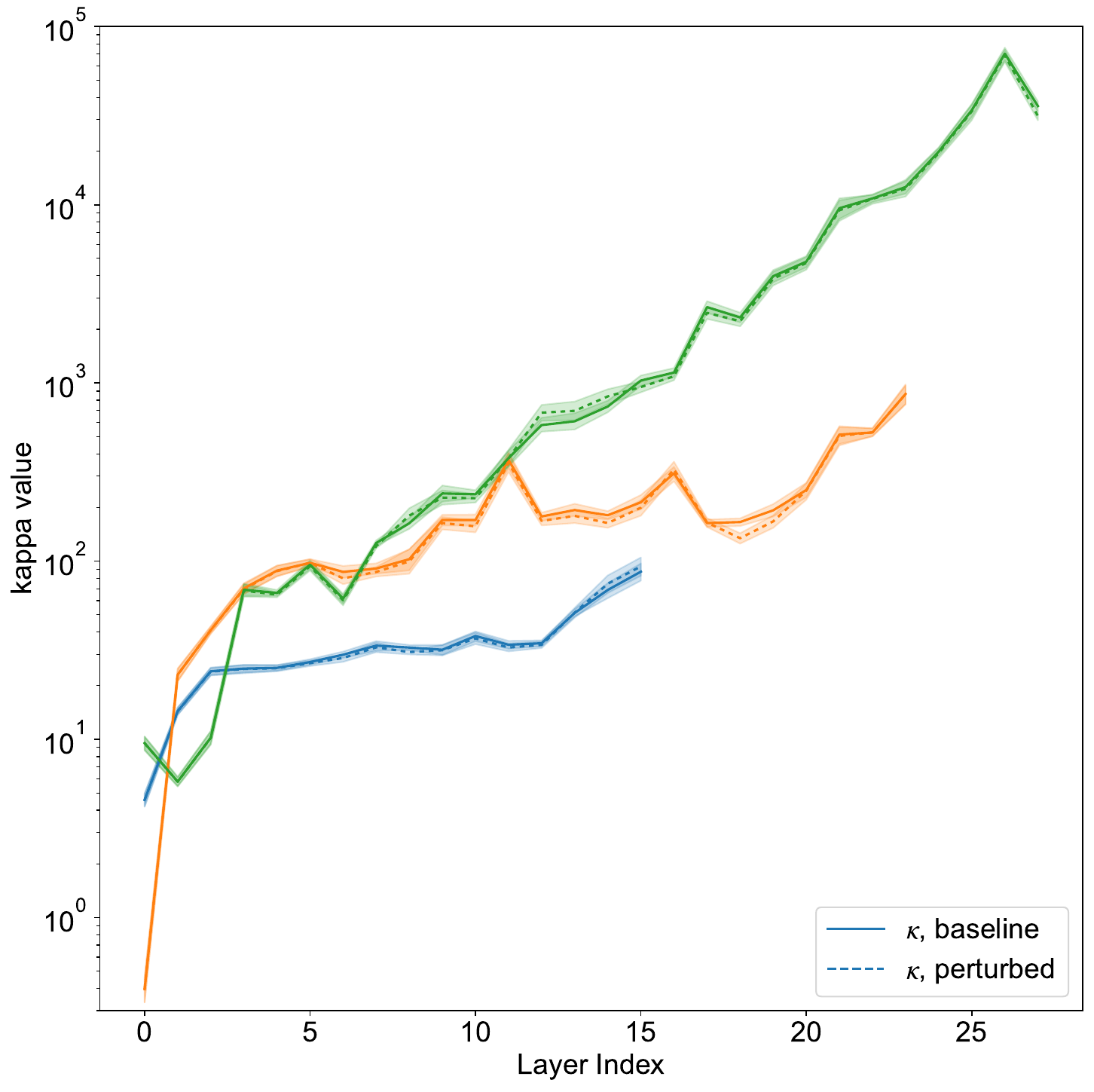}
  \caption{
    Layer‑wise vMF concentration $\kappa$ (log scale) for Llama3.2, Qwen2.5, and Qwen3 under baseline (solid) and perturbed (dashed) conditions.
    $\kappa$ determines $\mathrm{KL}(p\|U)$ and differs by orders of magnitude across models;
    perturbation leaves these orders unchanged, indicating that the sharpness of $p$ is unaffected by context shuffling.
  }
  \label{fig:kappa}
\end{figure}



\end{document}